\documentclass[10pt,twocolumn,letterpaper]{article}

\usepackage[pagenumbers]{cvpr} 
\usepackage{enumerate}
\usepackage{times}
\usepackage{epsfig}
\usepackage{graphicx}
\usepackage{amsmath}
\usepackage{amssymb}
\usepackage{multirow}
\usepackage{appendix}
\usepackage[dvipsnames]{xcolor}

\usepackage{xspace}
\usepackage{color}
\usepackage{adjustbox}
\usepackage{enumitem}
\usepackage{paralist} 
\usepackage{booktabs} 
\usepackage{array}
\usepackage[hypcap=true]{caption} 
\graphicspath{{figures/}}

\newcommand{\Paragraph}[1]{{\flushleft{\textbf{#1}}}} 

\newcommand{\mb}[1]{\mathbf{#1}}

\definecolor{gray}{rgb}{0.35,0.35,0.35}
\definecolor{MyBlue}{rgb}{0,0.2,0.8}
\definecolor{MyRed}{rgb}{0.8,0.2,0}
\definecolor{MyGreen}{rgb}{0.0,0.4,0.1}
\definecolor{MyGray}{rgb}{0.4,0.4,0.4}

\long\def\ignorethis#1{}

\newlength\paramargin
\newlength\figmargin
\newlength\secmargin

\newcolumntype{L}[1]{>{\raggedright\let\newline\\\arraybackslash\hspace{0pt}}m{#1}}
\newcolumntype{C}[1]{>{\centering\let\newline\\\arraybackslash\hspace{0pt}}m{#1}}
\newcolumntype{R}[1]{>{\raggedleft\let\newline\\\arraybackslash\hspace{0pt}}m{#1}}

\def\ie{\textit{i.e.},\xspace}
\def\eg{\textit{e.g.},\xspace}

\def\vs{vs.~}

\newcommand{\secref}[1]{Section~\ref{sec:#1}}
\newcommand{\figref}[1]{Fig.~\ref{fig:#1}}
\newcommand{\tabref}[1]{Table~\ref{tab:#1}}

\newcommand{\algref}[1]{Algorithm~\ref{alg:#1}}

\RequirePackage[numbers,sort&compress]{natbib}
\usepackage[ruled,linesnumbered,boxed]{algorithm2e}
\SetKwComment{Comment}{$\triangleright$\ }{}
\usepackage[pagebackref,breaklinks,colorlinks]{hyperref}

\title{MAG-Edit: Localized Image Editing in Complex Scenarios via 
\\
\underline{M}ask-Based \underline{A}ttention-Adjusted \underline{G}uidance}

\author{\vspace{0.2cm}Qi Mao$^{1}$ \hspace{0.65cm} Lan Chen$^1$ \hspace{0.65cm} Yuchao Gu$^2$ \hspace{0.65cm} Zhen Fang$^1$ \hspace{0.65cm} Mike Zheng Shou$^2$ \\
$^1$MIPG, Communication University of China \hspace{0.65cm} $^2$Show Lab, National University of Singapore \vspace{0.2cm} \\
\url{https://mag-edit.github.io/}}
\begin{document}
\maketitle
\begin{abstract}
Recent diffusion-based image editing approaches have exhibited impressive editing capabilities in images with simple compositions.
However, localized editing in complex scenarios has not been well-studied in the literature, despite its growing real-world demands.
Existing mask-based inpainting methods fall short of retaining the underlying structure within the edit region.
Meanwhile, mask-free attention-based methods often exhibit editing leakage and misalignment in more complex compositions.
In this work, we develop \textbf{MAG-Edit}, a training-free, inference-stage optimization method, which enables localized image editing in complex scenarios. 
In particular, MAG-Edit optimizes the noise latent feature in diffusion models by maximizing two mask-based cross-attention constraints of the edit token, which in turn gradually enhances the local alignment with the desired prompt.
Extensive quantitative and qualitative experiments demonstrate the effectiveness of our method in achieving both text alignment and structure preservation for localized editing within complex scenarios.
\end{abstract}    
\section{Introduction}
\label{sec:intro}
Text-based image editing aims to manipulate images in accordance with provided textual prompts.
Recent advancements in large-scale text-to-image (T2I) diffusion models, such as Stable Diffusion~\cite{rombach2022high}, DALL$\cdot$E~\cite{ramesh2022hierarchical}, and Imagen~\cite{saharia2022photorealistic}, have demonstrated remarkable ability to generate high-quality, diverse images that accurately reflect specified textual descriptions.
Trained on comprehensive datasets, these models effectively connect textual descriptions with corresponding images, thereby paving a new way for text-based image editing.

The past year has witnessed a substantial increase in the development of methods using diffusion models for text-based image editing, which can be broadly categorized into three groups:
training~\cite{brooks2023instructpix2pix,zhang2023magicbrush}, fine-tuning~\cite{kawar2023imagic,zhang2023sine,ruiz2023dreambooth}, and training-free methods~\cite{hertz2022prompt,tumanyan2023plug,parmar2023zero,mokady2023null,couairon2022diffedit,cao_2023_masactrl}.
Existing approaches predominantly concentrate on manipulating \emph{prominent} objects within \emph{simple} compositions.
However, images in real-world scenarios usually contain intricate compositions with multiple objects.
Additionally, users often require edits in specific localized regions.
For example, in home interior design, a user might wish to change the color of a particular piece of furniture to better complement the surrounding space. 
A case in point is altering the color of a sofa to green, as illustrated in the first row of \figref{teaser}, to improve its aesthetic coherence with the environment.

The trade-off between \emph{fidelity} and \emph{editability} in localized image editing within complex scenarios presents significant challenges.
Mask-based inpainting methods directly generate a new object as a foreground element and blend it into the original image~\cite{avrahami2022blended,avrahami2023blended,couairon2022diffedit,wang2023instructedit,huang2023pfb}.
However, this often results in substantial structural changes within the edited areas, causing noticeable discordance with their complex surroundings, as shown in the third column of \figref{teaser}.
On the other hand, mask-free methods that utilize attention injection mechanisms such as Prompt-to-Prompt (P2P)~\cite{hertz2022prompt} and Plug-and-Play (PnP)~\cite{tumanyan2023plug} can preserve the original image's structure and layout.
Nevertheless, they struggle to precisely align the local editing region with the intended text in intricate scenarios, largely due to their reliance on the text prompts' localization capabilities. As a result, editing effects often extend beyond the intended area and impact incorrect regions, as shown in the fourth column of \figref{teaser}.
Integrating mask-based blending techniques into P2P and PnP can alleviate leakage, but issues with incorrect alignment remain unresolved. This misalignment leads to the absence of editing effects in the intended areas, demonstrated in the last column of \figref{teaser}.

In this work, we introduce a novel editing scheme named \textit{\textbf{M}ask-Based \textbf{A}ttention-Adjusted \textbf{G}uidance} (\textbf{MAG}-Edit). This approach is designed to enable \emph{localized} image editing in \emph{complex} scenarios, which typically involve intricate compositions with multiple objects.
Given that cross-attention (CA) maps in pre-trained T2I diffusion models effectively capture the correlation between input features and text embeddings, our key insight is that \emph{adjusting the noise latent feature to attain higher CA values significantly enhances its alignment with the corresponding text prompt}.
As a result, we propose locally optimizing the noise latent feature during the inference stage by maximizing two distinct mask-based CA constraints tailored for the target editing prompt. 
In particular, our approach aims to maximize two aspects of ratios: first, the CA value of the edit token in relation to all token CA values within the masked area, and second, the CA value of the edit token inside the mask compared to its overall CA values.
Subsequently, the gradients of these constraints guide the update of the noise latent feature, thus progressively aligning the editing effect with the desired text prompt and spatial requirements. 
The effectiveness of the proposed method is evident in the second column of \figref{teaser}.

The main contributions of our work can be summarized as follows,

\begin{compactitem}
    \item 
    We introduce MAG-Edit, a novel training-free, inference-stage optimization scheme. 
    To our knowledge, this is the first method specifically designed to address localized image editing in complex scenarios.
    \item 
We propose two mask-based CA constraints in terms of the token and spatial ratio, guiding the local noise latent feature to better align with the target text.
    \item  
    We extensively validate MAG-Edit's efficiency in localized image editing across diverse intricate indoor and outdoor scenarios.
    Quantitative and qualitative experimental results demonstrate a significantly improved trade-off between editing efficiency and structure preservation when compared to existing baselines.
\end{compactitem}

\section{Related Work}
\label{sec:related}
\Paragraph{Text-Based Image Editing Using Diffusion Models} can be mainly classified into three categories:  training~\cite{brooks2023instructpix2pix,zhang2023magicbrush}, fine-tuning~\cite{kawar2023imagic,zhang2023sine,ruiz2023dreambooth}, and training-free methods~\cite{hertz2022prompt,tumanyan2023plug,parmar2023zero,mokady2023null,couairon2022diffedit,cao_2023_masactrl}.
InstructPix2Pix~\cite{brooks2023instructpix2pix} requires significant resources for extensive training, while fine-tuning methods like Imagic~\cite{kawar2023imagic} risk overfitting by optimizing the full model with limited data. 
In this work, we focus on training-free methods.
%
%
Some approaches~\cite{avrahami2022blended,avrahami2023blended,huang2023pfb,wang2023instructedit} utilize masks to generate foreground objects and blend them into the original image through blending operations.
In particular, the Blended Diffusion~\cite{avrahami2022blended} and Blended LD~\cite{avrahami2023blended}, directly generate foreground objects based on text prompts.
DiffEdit~\cite{couairon2022diffedit,wang2023instructedit} introduces an unsupervised method for learning the mask and employs DDIM inversion~\cite{song2020denoising} noise latent features alongside the target prompt to generate the foreground image.
Although these approaches successfully maintain the integrity of unedited regions outside of the mask, they may introduce large structural changes within the edit regions, causing inconsistencies with the surrounding context in complex scenes.
Other methods~\cite{hertz2022prompt,tumanyan2023plug,cao_2023_masactrl} such as P2P~\cite{hertz2022prompt} involve the attention integration mechanisms to maintain the structure and layout of the original image.
Recent advancements in inversion methods~\cite{li2023stylediffusion,mokady2023null,miyake2023negative} propose to improve DDIM inversion~\cite{song2020denoising} for encoding real images, achieving improved reconstruction and more flexible editing capabilities. However, the integration of P2P~\cite{hertz2022prompt} remains essential for these methods to facilitate image editing.
When applied to localized editing in intricate scenarios, attention-based methods often result in leakage to incorrect areas, leading to inefficiencies in prospective regions.

\Paragraph{Optimization on the Noise Latent Feature.}
Recent advances~\cite{chefer2023attend,chen2023training,xie2023boxdiff} in image generation with diffusion models have investigated the use of CA constraints to optimize the noise latent feature during inference.
The pioneering work, Attend-and-Excite~\cite{chefer2023attend},  addresses issues like catastrophic neglect and incorrect attribute binding by maximizing the largest CA units corresponding to all subject tokens in the text prompt. 
This approach refines the noise latent feature at each diffusion step, thereby guiding the model to generate all subjects described in the text accurately.
Several training-free layout-generation methods ~\cite{chen2023training,xie2023boxdiff} propose to optimize the noise latent feature by maximizing CA constraints in conjunction with bounding boxes, allowing objects to appear in specific regions.
While the image generation process has demonstrated effectiveness, the application of noise latent feature optimization to image editing has received relatively less attention. 
Pix2pix-zero~\cite{parmar2023zero} offers a solution by optimizing the noise latent feature, constraining the CA maps of the editing branch to align with the reconstruction branch, thus preserving the original image's structural layout.
In contrast to structural preservation, the proposed method aims to align the local noise latent feature more semantically with the target text prompt, enabling localized editing in complex scenarios.

\begin{figure}[!t]
    \centering
    \includegraphics[width=1\linewidth]{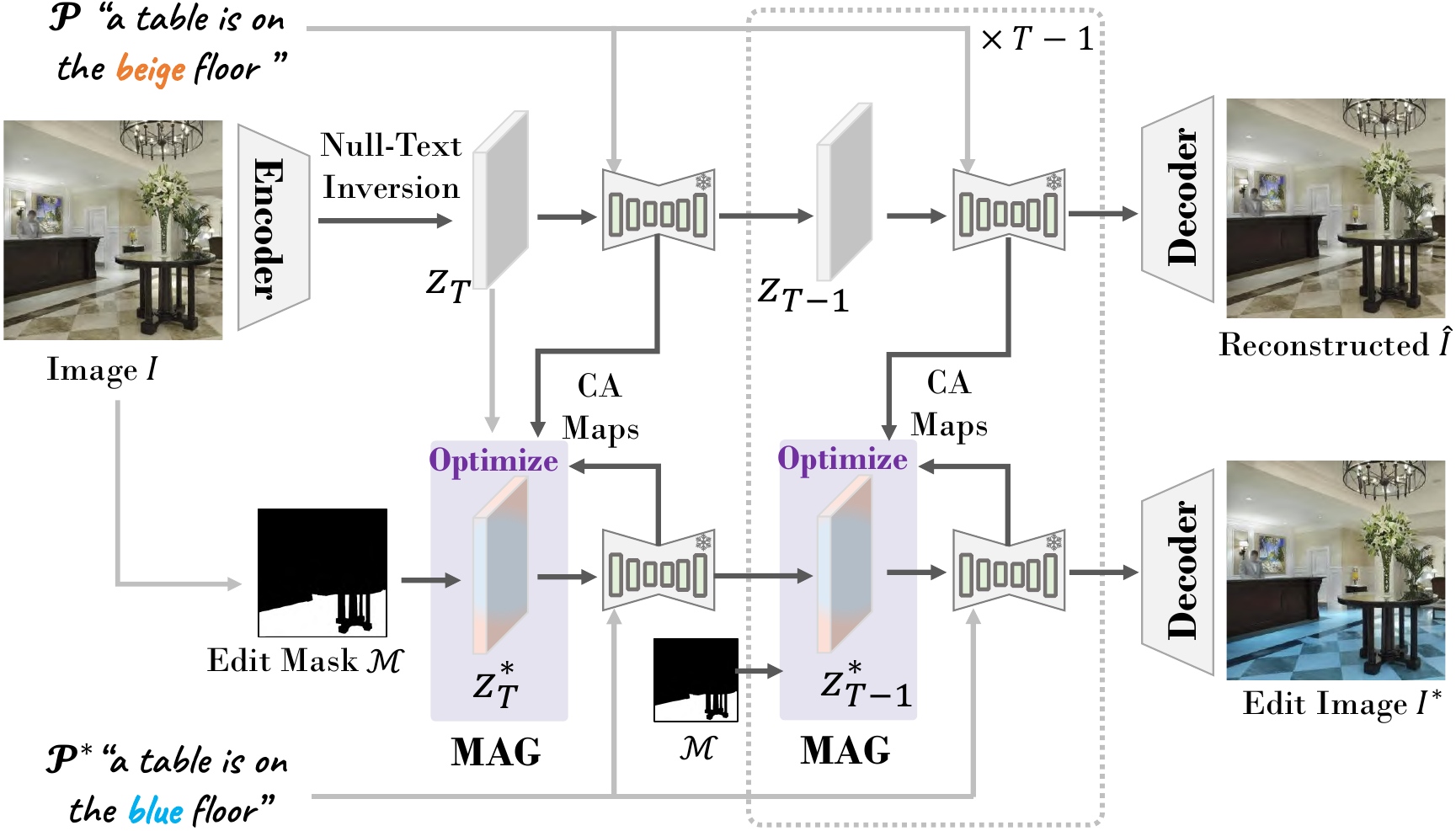}    \caption{\textbf{High-level overview of the proposed MAG-Edit framework.} 
    The first and second rows represent the reconstruction and editing branches, respectively.
    In the editing branch, the noise latent feature is optimized through MAG.
    This optimization process aids in achieving alignment with the target edit prompt ``{\color{cyan}{\textbf{blue}}}'' within the intended edit region $\mathcal{M}$.
    }
    \label{fig:framework}
    \vspace{-5 mm}
\end{figure}

\section{Background and Preliminaries}
\label{sec:pre}
\Paragraph{Stable Diffusion (SD)}~\cite{rombach2022high} aims to denoise the random sampled noise latent $z_T$ conditioned on text embedding $\mathcal{C}$.
This process transforms $z_T$ into a series of noise latent features $z_t$ at each diffusion step $t$, where $t\sim \left[ 1,T \right]$ and $T$ is the timestep number.
To train the diffusion model $\varepsilon_\theta$, the initial latent feature $z_0$ undergoes an interactive process by adding Gaussian noise $\varepsilon$ to the noise latent features $z_t$.
Then, the network is minimized by,
\begin{equation}
\min _\theta E_{z_0, \varepsilon \sim N(0, I), t \sim \text { Uniform }(1, T)}\left\|\varepsilon-\varepsilon_\theta\left(z_t, t, \mathcal{C}\right)\right\|_2^2.
\end{equation}
Furthermore, the classifier-free guidance~\cite {ho2022classifier} performs unconditional prediction to  mitigate the amplifying effect of text-based conditioning as:
\begin{equation}
\small
\tilde{\varepsilon}_\theta\left(z_t, t, \mathcal{C}, \varnothing\right)=w \cdot \varepsilon_\theta\left(z_t, t, \mathcal{C}\right)+(1-w) \cdot \varepsilon_\theta\left(z_t, t, \varnothing\right),
\end{equation}
where $\varnothing$ is the unconditional embedding of a null text, and $w$ is the guidance weight.
To generate images from given $z_T$, we can employ deterministic DDIM sampling~\citep{song2020denoising} as:
\begin{equation}
\scriptsize
z_{t-1}=\sqrt{\frac{\alpha_{t-1}}{\alpha_t}} z_t+\left(\sqrt{\frac{1}{\alpha_{t-1}}-1}-\sqrt{\frac{1}{\alpha_t}-1}\right) \cdot \tilde{\varepsilon}_\theta\left(z_t, t, \mathcal{C}, \varnothing\right).
\end{equation}
%

\begin{figure}[!t]
    \centering
    \includegraphics[width=0.92\linewidth]{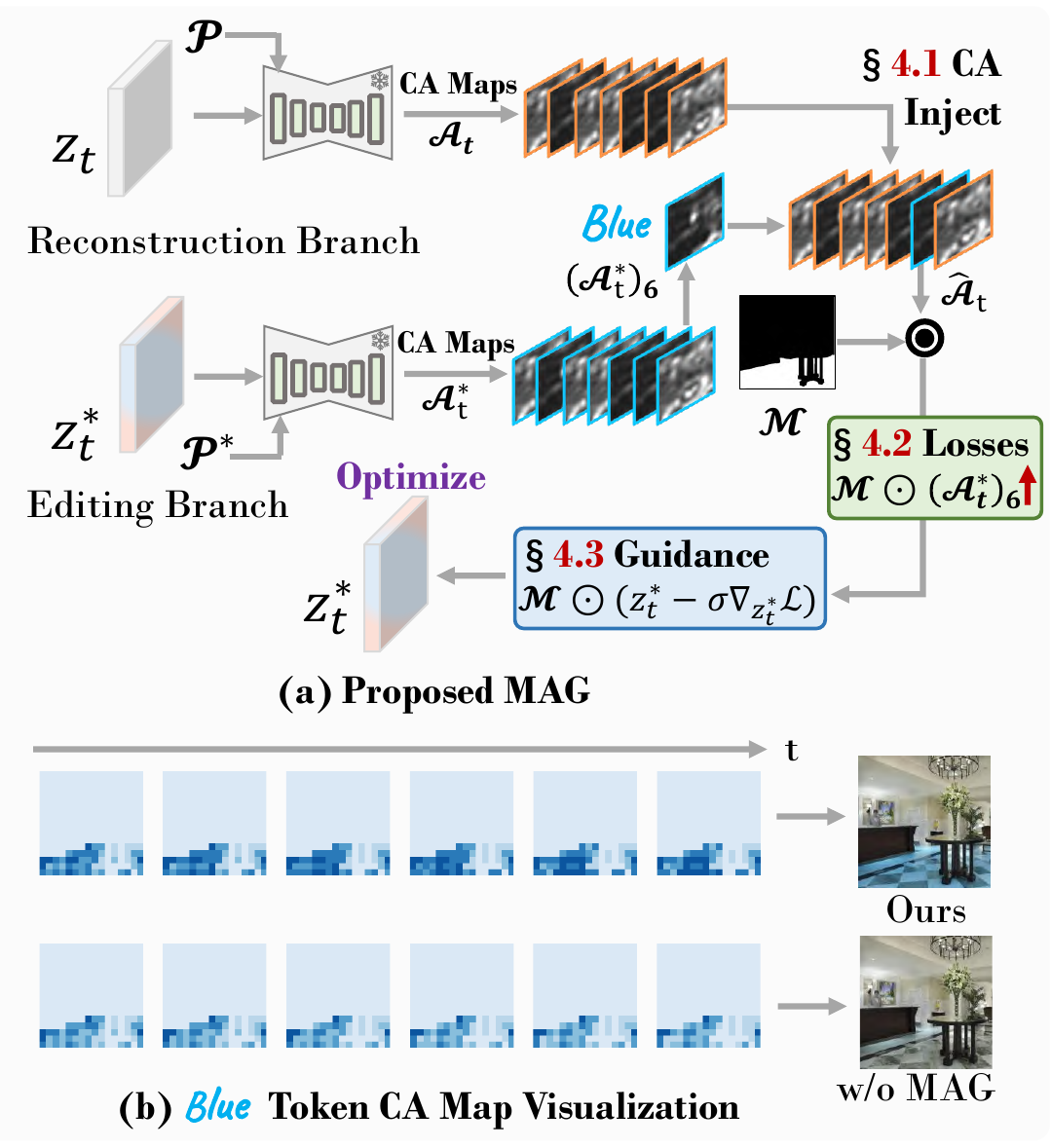}    \caption{\textbf{Illustration of Our MAG.} 
    (a) We optimize the $z^*_t$ by maximizing the mask-based CA constraints of target token (\eg ``{\color{cyan}{\textbf{blue}}}''). 
    (b) The top row and the bottom row illustrate the average mask-based CA maps of ``{\color{cyan}{\textbf{blue}}}'' token with our MAG and ``w/o MAG'' from different diffusion steps, respectively.
    Applying optimization of $z^*_t$ with MAG,  there is a noticeable enhancement in the CA values within the edit regions $\mathcal{M}$.
    }
    \label{fig:MAG}
    \vspace{-3 mm}
\end{figure}

\Paragraph{Null-Text Inversion.}
Real image editing requires reversing corresponding $z_{0}$ back to $z_{T}$.
A straightforward DDIM inversion method~\citep{song2020denoising}, in theory reversible with infinitesimally small steps, tends to accumulate reconstruction errors in the denoising process, particularly due to classifier guidance. To address this, Null-text inversion \cite{mokady2023null} aligns the diffusion latent trajectory with the denoising latent trajectory by optimizing a step-wise unconditional embedding $\varnothing_t$.

\Paragraph{Prompt-to-Prompt}
(P2P)~\cite{hertz2022prompt} introduces several prompt-based editing operations leveraging CA maps: 
First, the \emph{word swap} involves injecting all attention maps in the reconstruction branch, generated by the source prompt, into attention maps from the editing process using the target prompt.
In contrast, the \textit{prompt refinement} selectively replaces the CA maps associated with tokens common to both the source and target prompts.
Furthermore, P2P introduces the \emph{attention re-weighting} operation, enabling direct scale adjustments to the CA values.
This technique is specifically designed to control the granularity of the editing process.

\section{Methodologies}
Let $\mathcal{I}$ be a real image, we first employ Null-text inversion~\cite{mokady2023null} to encode it
into the noise latent feature $z_{T}$.
Given the original text prompt $\mathcal{P}$ and edited prompt $\mathcal{P^\ast}$, we define the set of new target tokens as $\mathcal{S}^\ast=\{s^*_1, s^*_i,...,s^*_I\}$ present in $\mathcal{P^*}$ against $\mathcal{P}$, the common tokens as $\mathcal{S}=\{s_1, s_j,...,s_J\}$ and $\mathcal{S}^\ast \cap \mathcal{S} = \emptyset$.
An edit region mask $\mathcal{M}$ derived by $\mathcal{I}$ is provided to precisely localize the edit region.
\figref{framework} illustrates the high-level overview of the proposed editing framework, which consists of two branches, \ie reconstruction and editing branches generated by prompt $\mathcal{P}$ and $\mathcal{P}^*$, respectively.
In this work, we aim to optimize the noise latent feature $z^\ast_{t}$ of the editing branch at diffusion step $t$.
Our objective is to align the desired editing effects specified by $\mathcal{S}^\ast$ with the prospective region defined by $\mathcal{M}$, which enables localized editing in complex scenarios.

To achieve this, we first inject CA maps of common tokens similar to the \emph{prompt refinement} in P2P~\cite{hertz2022prompt} (\secref{atten}).
Subsequently, we introduce MAG to automatically manipulate $z^{\ast}_{t}$, which contains two key steps: defining two mask-based constraints in \secref{constraint} and performing gradient guidance in \secref{gradient}.

\subsection{Attention Injection}
\label{sec:atten}
As illustrated in~\figref{MAG} (a), to preserve the structural information of the original image,
CA maps of common tokens from the reconstruction branch are first injected into the editing branch at diffusion step $t$, thereby obtaining the mixing CA maps $\hat{A}_t$ as:
\begin{equation}
\footnotesize
   Inject(A_t,A^*_t) :=
   \begin{cases}
      (A_t)_{j}&  j \in \{1,j,\cdots,J\},\\
      (A^*_t)_{i}& i \in \{1,i,\cdots,I\}.\\
    \end{cases}
\label{Eq:proposed1}  
\end{equation}

\begin{figure*}[!t]
    \centering
    \includegraphics[width=0.995\linewidth]{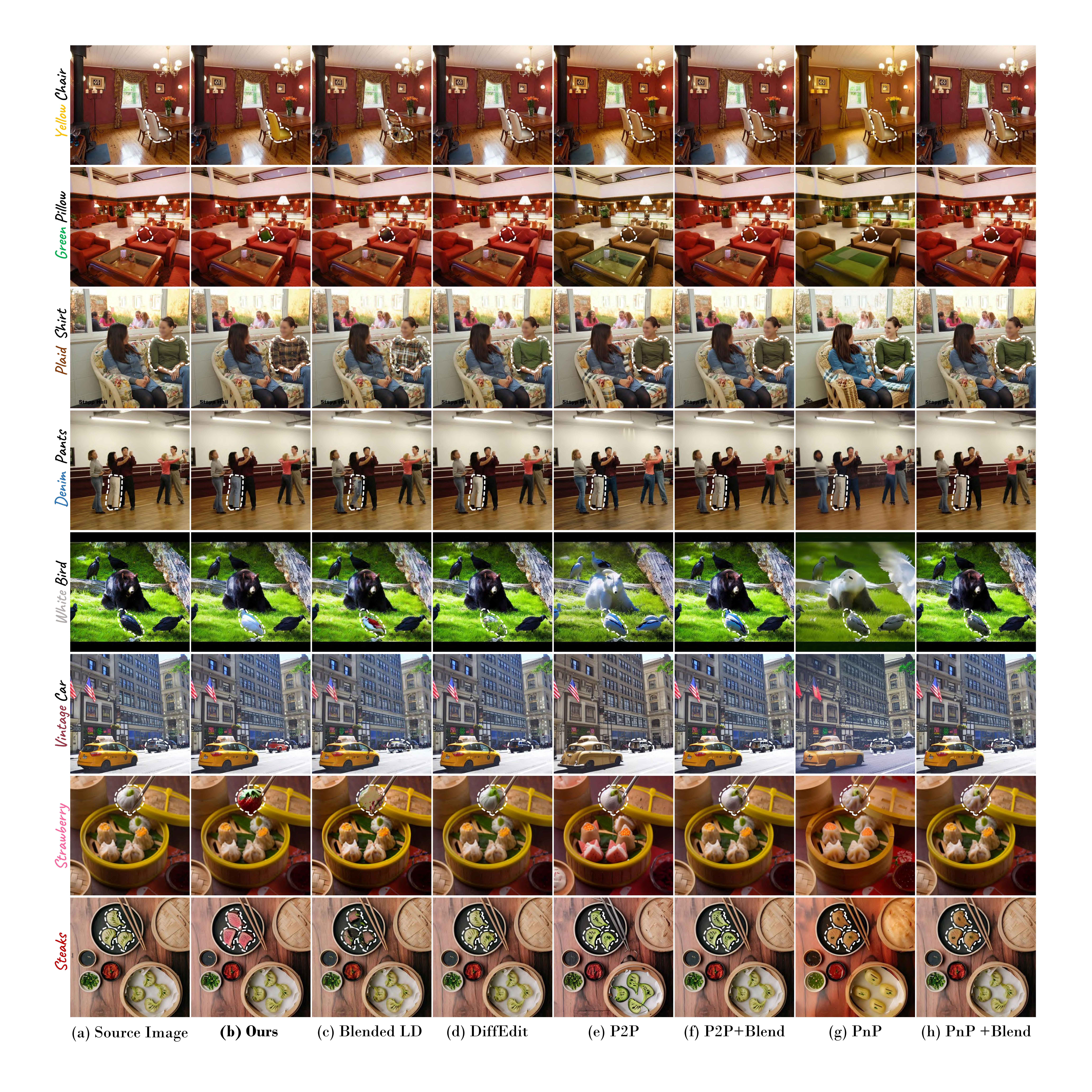}    \caption{\textbf{Qualitative comparisons of localized image editing across various complex scenarios.}
We highlight the editing regions with white dashed lines. 
Simplified target edit prompts are denoted on the left side of (a) source images. Our proposed method (b) not only achieves superior editing effects but also better preserves the structure in local regions
against other baselines (c-h).
    }
    \vspace{-5 mm}
    
    \label{fig:qualitative}
\end{figure*}

\subsection{Mask-Based Attention-Adjusted Constraints}
\label{sec:constraint}
Considering that CA maps define the similarity between the input features and text embeddings, larger CA values indicate better alignment.
This observation inspires the formulation of two mask-based constraints, aiming to maximize the CA value ratio in both token and spatial aspects within the predefined editing region.
To illustrate, first consider the CA map $(A^\ast_t)_i$ of a new editing token $ s^\ast_i$ within a specific mask region $\mathcal{M}$ such as ``{\color{cyan}{\textbf{blue}}}'' in ~\figref{MAG} (a).
\Paragraph{Token Ratio Constraint.} 
Since CA maps of common tokens from the reconstruction process are first injected into the editing branch, this leads to the CA value of the new token $(A^\ast_t)_i$  being comparatively lower in contrast to other common tokens.
We then introduce a token ratio constraint that prioritizes increasing the value proportion of the new token within mask $\mathcal{M}$ among all tokens:
\begin{equation}
\footnotesize
\mathcal{L}_{TR}=\left (1-\frac{1}{\mathcal{\overline{M}}}\sum \mathcal{M} \odot  \frac{(A^\ast_t)_i}{(A^\ast_t)_i  +\sum_{j=1}^{J}(A_t)_j}\right )^2,
\label{Eq:proposed2}  
\end{equation}
where $\mathcal{\overline{M}}$ represents the total number of elements within the mask.

\Paragraph{Spatial Ratio Constraint.} 
In scenarios demanding significant editing granularity, the token ratio constraint might not sufficiently amplify the CA value $(A^\ast_t)_i$ within $\mathcal{M}$.
To address this limitation, we introduce an additional spatial formulation, which is designed to maximize the CA values within the masked region while simultaneously minimizing them outside the mask as,
\begin{equation}
\footnotesize
\mathcal{L}_{SR}=\lambda \underbrace{\left(1-\frac{\sum \mathcal{M} \odot (A^\ast_t)_i}{\sum (A^\ast_t)_i}\right)}_{\text{Out-mask}} - \underbrace{\frac{\sum \mathcal{M} \odot (A^\ast_t)_i}{\sum (A^\ast_t)_i}}_{\text{In-mask}},
\label{Eq:proposed3}
\end{equation}
where $\lambda$ is a balance weight, and we set $\lambda=3$ empirically.
%
%

\begin{algorithm}[!t]
\caption{A Denoising Step using MAG-Edit}
\label{alg:update}
\SetAlgoLined
\KwIn{A original and edited prompt $\mathcal{P}$, $\mathcal{P}^*$; a timestep $t$ and corresponding noise latent features of reconstruction and editing branches $z_t$, $z_t^*$; a maximum iteration step MAX\_IT; a function $\mb{F}(\cdot)$ for computing proposed constraints;a pre-trained Stable Diffusion model $SD$.}
\KwOut {the noisy latent feature $z^*_{t-1}$ for the next timestep of the editing branch.}
\For{$i=1$ \KwTo $\rm{MAX\_IT}$}
{
$\_, A_t \gets SD(z_t,\mathcal{P},t)$ \;
$\_, A^*_t \gets SD(z^*_t,\mathcal{P}^*,t)$ \;
$\hat{A}_t \gets Inject(A_t,A^*_t)$ \; 
            $\mathcal{L} \gets \mb{F}(\hat{A}_t) $\;
        $z_t^*=\mathcal{M}\odot(z_t^* - \delta \nabla_{z_t^*}\mathcal{L}) 
    + (1-\mathcal{M}) \odot z^*_t$\;
}
$\_, A_t \gets SD(z_t,\mathcal{P},t)$ \;
$\_, A^*_t \gets SD(z^*_t,\mathcal{P}^*,t)$ \;
$\hat{A}_t \gets Inject(A_t,A^*_t)$ \; 
$z^*_{t-1} \gets SD(z^*_t, \mathcal{P}^*,t)\{\hat{A}_t\}$ \;
\textbf{Return} $z^*_{t-1}$
\end{algorithm}

\Paragraph{Negative Prompt Constraint.}
In real image editing, the latent noise feature $z_{T}$ derived by the inversion methods still retains information related to the original image $\mathcal{I}$. 
Achieving the desired editing results can be challenging in some cases when there is a significant difference between the texture in the original image and modified prompt $\mathcal{P}^*$, such as transferring color from ``black" to ``white".
Our proposed method can also be used to attenuate the textural information associated with the original image $\mathcal{I}$ by employing negative prompts.
In particular, we define a set of negative tokens $\mathcal{S}^\ast_{\rm{ng}}$ to present the texture of $\mathcal{I}$ in contrast to the new tokens $\mathcal{S}^\ast$.
For example, if $\mathcal{P}^*$ is ``a man wears a white T-shirt" and the T-shirt in $\mathcal{I}$ is black, then the negative token would be ``black".
Consequently, we can establish the negative prompt constraint $\mathcal{L}_{\rm{ng}}$ using the negative token's corresponding CA value and optimize the noise latent feature in the opposite direction as follows,
%
\begin{equation}
\mathcal{L}_{\rm{total}}=\lambda_{\rm{p}}\mathcal{L}- \lambda_{\rm{ng}}\mathcal{L}_{\rm{ng}},
\label{Eq:proposed4}  
\end{equation}
where $\lambda_{\rm{p}}$ and $\lambda_{\rm{ng}}$ aim to balance between positive and negative prompt constraint.

\subsection{Perform Gradient Guidance}
\label{sec:gradient}
Upon establishing the mask-based constraints, we compute their gradients to determine the optimal direction for modifying the current noise latent feature $z^*_t$.
In particular, to restrict the editing effect to the predefined region, we update the noise latent feature $z^\ast_t$ inside the mask $\mathcal{M}$ using the following equation:
\begin{equation}
  z_t^*=\mathcal{M} \odot (z_t^* - \delta \nabla_{z_t^*}\mathcal{L}) + (1-\mathcal{M}) \odot z_t^*,
\label{Eq:proposed5}  
\end{equation}
where the term $\delta$ represents the gradient update scale.
As detailed in \algref{update}, $z_t^*$ is iteratively refined until reaching the maximum number of iteration.

Moreover, our proposed method can be readily adapted for multiple prompt editing as:
\begin{equation}
\small
\begin{split}
      z_t^*=\mathcal{M} \odot (z_t^* - \delta \nabla_{z_t^*} \sum_{i=1}^{I}(  \lambda_1\mathcal{L}^1 & + \cdots + \\
    \lambda_i\mathcal{L}^i  +\lambda_I\mathcal{L}^I))  
      & + (1-\mathcal{M}) \odot z_t^*,
\label{Eq:proposed6}  
\end{split}
\end{equation}
where the term $\lambda_{\ast}$ controls the editing granularity of each prompt, with their sum equaling $1$.
\figref{application} (a) demonstrates how our proposed method effectively balances the editing granularity for various prompts.
%

\begin{table}[t!]
    \centering
    \resizebox{\linewidth}{!}{
    \begin{tabular}{l cc ccc}
        \toprule 
        \multirow{3}{*}{\textbf{Method}} 
        & \multicolumn{2}{c}{\textbf{Quantitative Metrics} }
        & \multicolumn{3}{c}{\textbf{Human Preference (Ours \vs)}  }
       \\

        \cmidrule(lr){2-3} \cmidrule(lr){4-6} 

& \multirow{2}{*}{\shortstack[c]{CLIP\\ Score ($\uparrow$) }}  
& \multirow{2}{*}{\shortstack[c]{DINO-ViT \\ Distance ($\downarrow$) }} 
        
 & \multirow{2}{*}{\shortstack[c]{\\ Text\\ Alignment ($\%$) }}  
  & \multirow{2}{*}{\shortstack[c]{\\ Structure\\ Preservation ($\%$) }}  
 & \multirow{2}{*}{\shortstack[c]{Overall \\ Preference ($\%$) }} 
\\ \\
        \cmidrule(lr){1-6}

        Blended LD~\cite{avrahami2023blended}
        & 19.12 & 0.089 
        & \underline{84 \%}
        & \textbf{75\%}
        & \underline{80\%} \\
     
        Diffedit~\cite{couairon2022diffedit}
        & 19.20 & 0.083 
        & 77 \%
        & \underline{66\%}
        & 71 \%\\

        P2P \cite{hertz2022prompt} 
        & \underline{20.02} & \textbf{0.079} 
         & \textbf{87 \%}
          & 54 \%

        & \textbf{81\%} \\

        PnP~\cite{tumanyan2023plug}
        & 19.90 & 0.083  
        & \textbf{87 \%}
         & 59 \%
          & 79 \%\\

        P2P+Blend
        & 19.77 & 0.081  & 83 \%
         & 62 \%
        & 73\% \\

        PnP+Blend
        & 19.47 & \underline{0.080} & 82 \%
         & 56 \%
        & 69\% \\

        \textbf{Ours}
        & \textbf{21.79} & 0.081 & /
        & /&/\\
        \bottomrule 
    \end{tabular}
    }
    \caption{
    \textbf{Quantitative comparisons of localized image editing.} 
    We assess all the metrics and human preferences in the \textbf{localized editing regions}. 
    ``Ours \vs'' indicates the proportion of users who favor our proposed method over the comparative approach.
    }
    \vspace{-2 mm}
    \label{tab:cmp_baselines}
\end{table}


\begin{figure}[!t]
    \centering \includegraphics[width=1\linewidth]{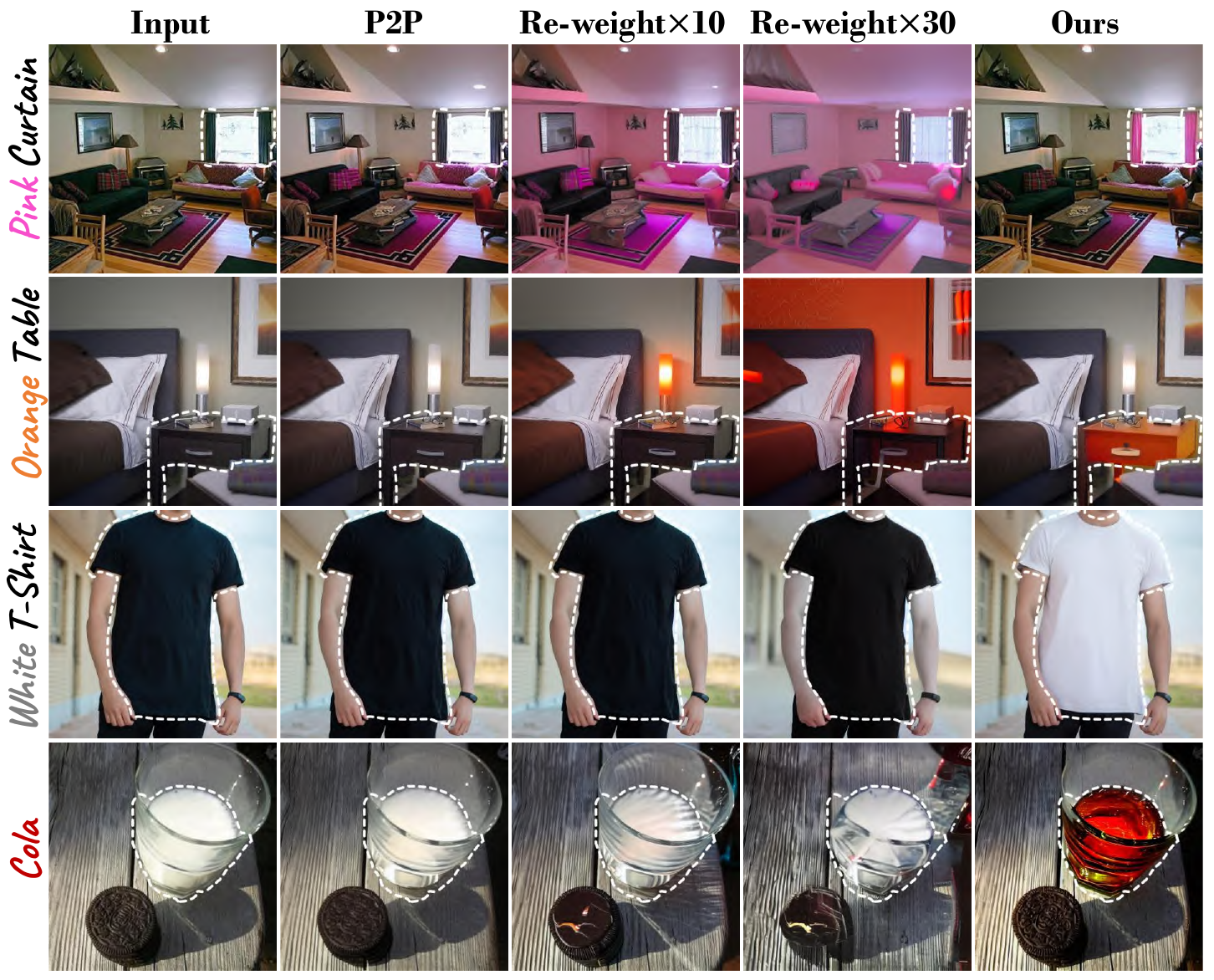}    \caption{\textbf{Attention re-weighting~\cite{hertz2022prompt}  \vs Ours.} 
   \emph{Attention re-weighting}~\cite{hertz2022prompt} either amplifies the entire editing magnitude in incorrect regions (first two rows) or fails to edit regions that significantly contradict the target prompt (last two rows).
In contrast, our proposed method effectively addresses both scenarios.
    }
    \vspace{-5 mm}
    \label{fig:ablation-reweight}
\end{figure}

\section{Experiments}
\subsection{Implementation details} 
We adopt the pre-trained Stable Diffusion v1.4~\cite{rombach2022high} model as the backbone.
All CA values are calculated in the resolution of $16 \times 16$ of the U-Net, which is known to process the most semantically rich information~\cite{chefer2023attend}. 
To preserve the original information in the regions outside the mask, we incorporate latent blend operation in~\cite{hertz2022prompt}.
In practice, we select between $\mathcal{L}_{TR}$ and $\mathcal{L}_{SR}$, depending on the required granularity of the edit, allowing for adaptability across various editing types.
All experiments are conducted on a single NVIDIA A100 GPU.
Additional implementation details are provided in ~\Cref{sec:implementation}.
\begin{figure}[!t]
    \centering
    \includegraphics[width=0.8\linewidth]{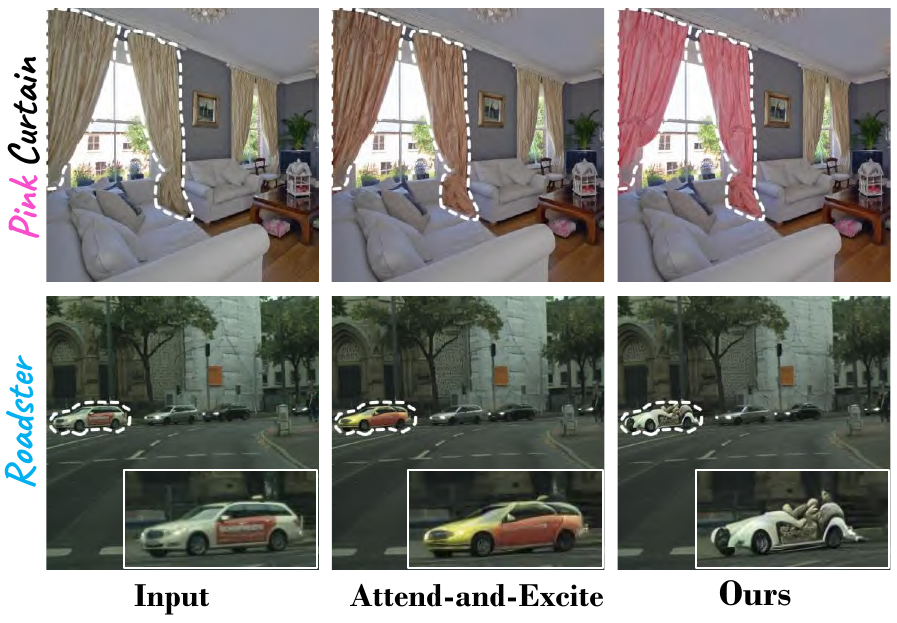}    \caption{\textbf{Attend-and-Excite~\cite{chefer2023attend} \vs Ours.}     
Attend-and-Excite's CA constraint~\cite{chefer2023attend} yields unsatisfactory editing results.
    }
    \label{fig:attend-and-excite}
    
\vspace{-5 mm}
\end{figure}

\subsection{Comparisons with Baselines}
\label{sec:baselines}
\Paragraph{Benchmark Dataset.} 
Existing datasets for text-based image editing methods primarily focused on relatively simple scenes dominated by prominent objects. 
To enable a more comprehensive evaluation of our method, we have curated a benchmark data set consisting of 200 images sourced from ADE20K~\cite{zhou2017scene}, MS-COCO~\cite{lin2014microsoft}, Cityscape~\cite{cordts2016cityscapes}, and the Internet. 
The selected images feature complex compositions with multiple objects in a wide range of real-world indoor and outdoor scenes.
Our evaluation primarily targets localized editing of color, texture, and object replacement.
We generate the source and target prompts using GPT-4~\cite{OpenAI2023GPT4TR}.
The corresponding edit masks are obtained using the Segment Anything method\footnote{\url{https://github.com/facebookresearch/segment-anything}}. 
Consequently, each image in the dataset is associated with three annotations: a source image prompt, a target image prompt, and the editing mask.
Additional details are provided in~\Cref{sec:benchmark}.

\Paragraph{Baselines.} 
We conduct comparisons with existing representative training-free diffusion-based image editing methods, covering these categories:
\begin{compactitem}
\item \textbf{Mask-based}: Blended Latent Diffusion (Blended LD)~\cite{avrahami2023blended} and DiffEdit~\cite{couairon2022diffedit}.
\item \textbf{Mask-free}: P2P~\cite{hertz2022prompt} and PnP~\cite{tumanyan2023plug}.
\item \textbf{Mask-free with blending}: We combine blending operations with P2P and PnP as baselines, denoted as P2P$+$Blend and PnP$+$Blend, respectively. 
Note that P2P$+$Blend can be viewed as the proposed method w/o MAG.
\end{compactitem}
%
%

\begin{figure}[!t]
    \centering
    \includegraphics[width=1\linewidth]{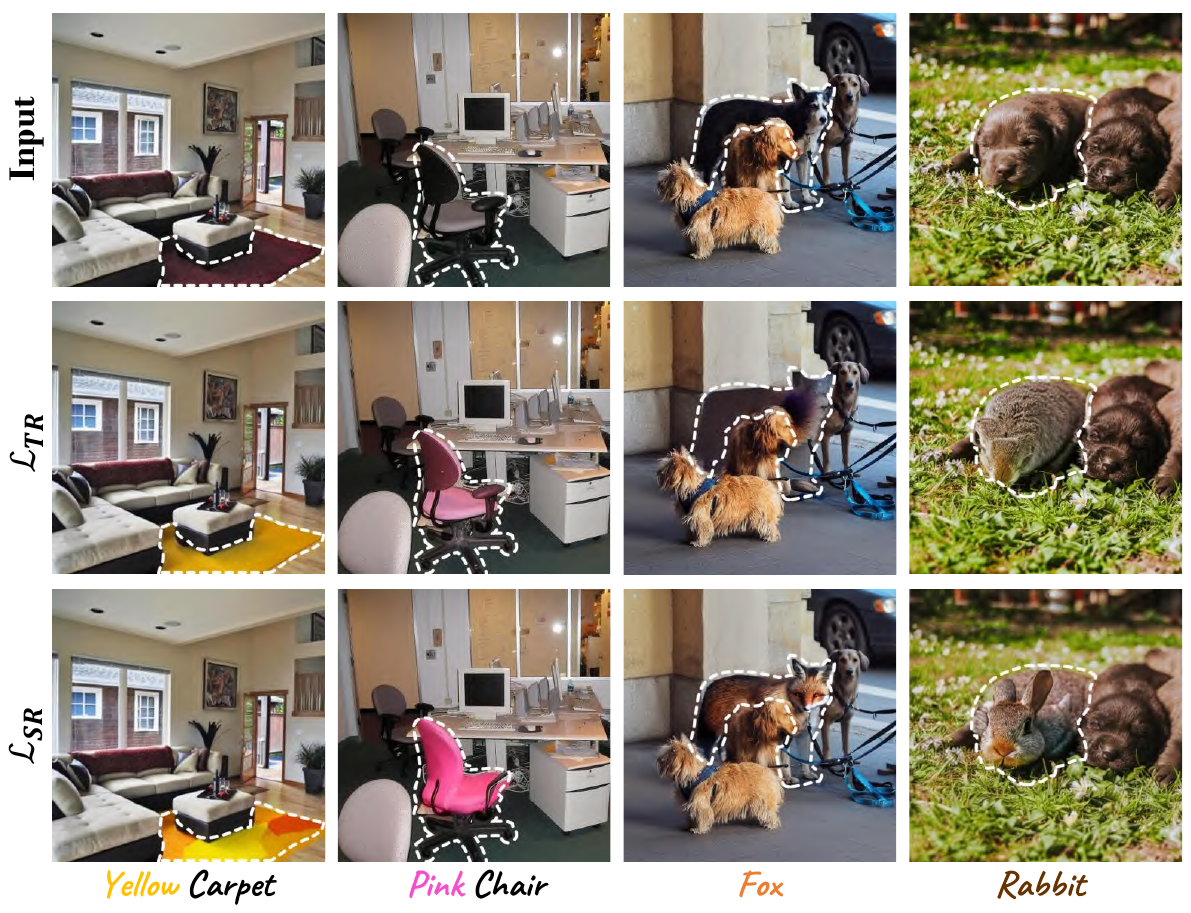}    \caption{\textbf{Editing granularity of proposed constraints.}  
     The token ratio constraint $\mathcal{L}_{TR}$ 
     efficiently preserves the inherent structure in the edited region, while the spatial ratio constraint $\mathcal{L}_{SR}$ enhances editing granularity.
    }
    \label{fig:losses}
\vspace{-5 mm}
\end{figure}

\Paragraph{Qualitative results.}
\figref{qualitative} clearly shows that Blended LD~\cite{avrahami2023blended} leads to considerable structural changes in complex scenarios, resulting in significant discordance with the surrounding context. 
Meanwhile, DiffEdit~\cite{couairon2022diffedit}, which employs DDIM inversion for foreground generation, either alters the structure, as seen in the ``white bird'' example in the fifth row, or fails to produce a noticeable editing effect in the intended region, as is apparent in other images.
Concerning mask-free methods, P2P~\cite{hertz2022prompt} and PnP~\cite{tumanyan2023plug} often exhibit leakage into adjacent regions, leading to minimal effects in the prospective region.
This issue is particularly noticeable in tasks such as changing the color of a yellow chair or a green pillow.
Blending operations might reduce leakage in some scenarios, yet the issue of misalignment persists, resulting in inefficiencies in the intended edit regions.
In contrast, our proposed method shows improved editing performance with better structural preservation.

\begin{figure*}[!t]
    \centering
    \includegraphics[width=0.995\linewidth]{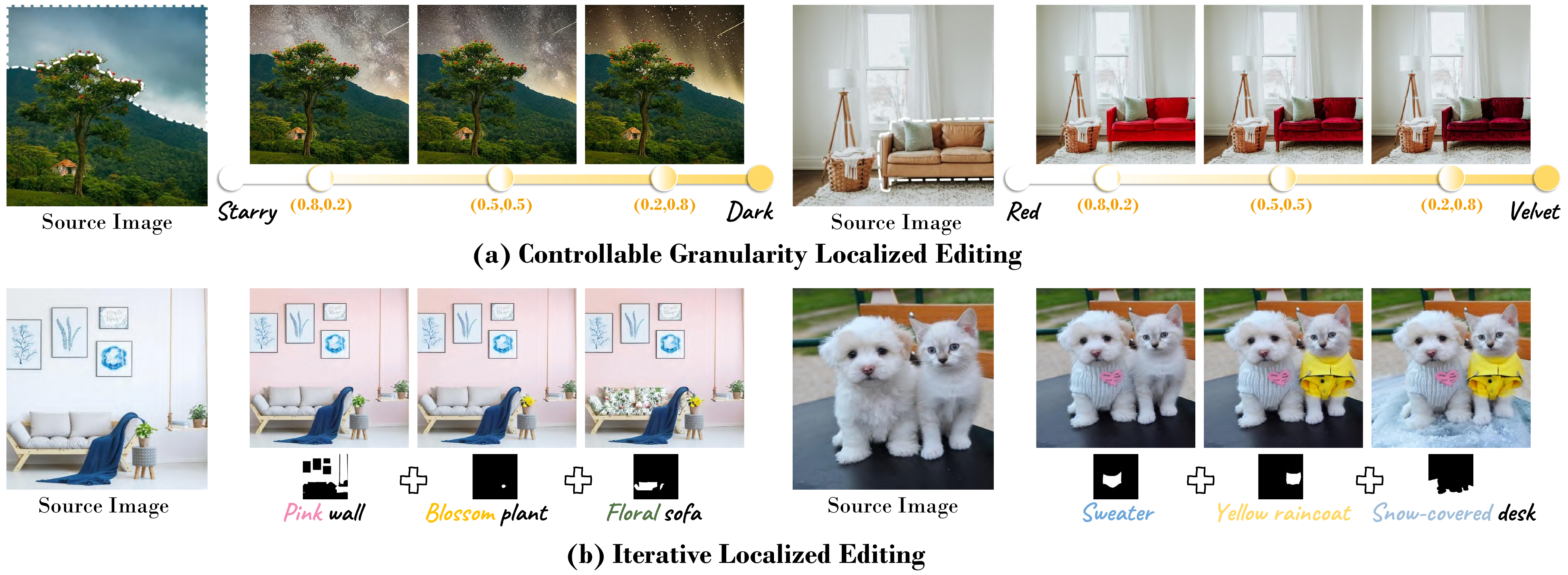}    \caption{\textbf{Other localized editing applications of the proposed MAG-Edit.}
    }
    \label{fig:application}
    \vspace{-5 mm}
\end{figure*}

\Paragraph{Quantitative results.}
We quantitatively evaluate our proposed method against baseline models using both automatic metrics and human evaluations. 

\emph{Automatic Metrics.} To better evaluate localized editing ability, we use the bounding boxes to crop the editing regions~\cite{huang2023pfb} and evaluate the image-text alignment and structure preservation using the CLIP score ~\cite{radford2021learning} and the DINO-ViT self-similarity distance~\cite{tumanyan2022splicing}, respectively.
\tabref{cmp_baselines} illustrates that our proposed method significantly enhances text alignment within local regions, achieving much higher local CLIP values without compromising fidelity.

\emph{User study.} 
We perform a user preference evaluation via pairwise comparisons on Amazon MTurk\footnote{\url{https://www.mturk.com/}}, focusing on text alignment, structure preservation, and overall preference in localized editing regions.
%
%
As shown in \tabref{cmp_baselines}, the percentages represent the proportion of users who prefer our proposed method over comparative approaches. 
A significant majority, ranging from $77\%$ to $87\%$, believe that our method achieves much better text alignment compared to other methods.
Furthermore, our method is preferred for better structure preservation by $75\%$ of users over Blended LD~\cite{avrahami2023blended}. 
Due to its more effective balance between editability and fidelity, our proposed method is overall favored by $69\%$ to $80\%$ of the participants.
%

\subsection{Ablation Study}

\Paragraph{Why Optimize the $z_t$.}
To enhance the editing effect, a straightforward approach is to directly increase the CA values of the corresponding token using the \emph{attention re-weighting} in P2P~\cite{hertz2022prompt}.
However, due to P2P's inherent misalignment, this direct amplification of CA values tends to intensify the editing effects in incorrect regions, failing to enhance the desired localized areas (first two rows in \figref{ablation-reweight}).
Additionally, minimizing the influence of information from the original image that conflicts with the edit prompt poses a challenge, even in prominent objects, such as changing color from ``black'' to ``white'' (last two rows in~\figref{ablation-reweight}).  
In contrast, our proposed method focuses on local alignment in specific regions and effectively attenuates contradicting information by directly optimizing the noise latent feature. 

\Paragraph{\vs Attend-and-Excite.}
Attend-and-Excite~\cite{chefer2023attend} optimizes the noise latent feature $z_t$ in the unconditional generation to maximize the largest CA value of the subject token.
We then establish a baseline using the CA constraint formulation in~\cite{chefer2023attend}. 
However, \figref{attend-and-excite} shows that this constraint is insufficient for image editing scenarios. 
Contrary to unconditional generation, the noise latent feature derived from inversion methods contains more information related to the real image, as opposed to random noise features sampled from a Gaussian distribution.
As such, our proposed constraints offer more efficient guidance for adjusting the noise latent feature, thereby more aptly addressing the needs of image editing.

\Paragraph{Impact of Proposed Constraints.} 
$\mathcal{L}_{TR}$ and $\mathcal{L}_{SR}$ offer distinct levels of editing granularity, as demonstrated in~\figref{losses}.
$\mathcal{L}_{TR}$ excels in maintaining the inherent structure within the edit region, which aids in achieving natural color and texture modifications.
On the other hand, $\mathcal{L}_{SR}$ provides stronger guidance by directly amplifying the CA values within the mask, leading to more noticeable structural changes in the edit region.
As a result, $\mathcal{L}_{SR}$ is better suited for edits involving large structural shape changes.


\subsection{Other Applications}
MAG-Edit is also adaptable for controllable granularity and iterative localized editing.
In~\figref{application} (a), MAG-Edit demonstrates the ability to balance editing granularity across various prompts, catering to user-specific requirements.
Furthermore, \figref{application} (b) illustrates MAG-Edit's capability to execute iterative, localized manipulations on various objects within a single image.
More results are shown in~\Cref{sec:results}.

\section{Conclusions}
In this work, we introduce a novel technique \emph{Mask-Based Attention-Adjusted Guidance} (MAG-Edit), specifically crafted for localized editing in complex scenarios.
In particular, we propose to maximize two mask-based CA constraints, namely token and spatial ratio, to locally optimize the noise latent feature for enhanced alignment with the target text embedding.
Our experimental results, both quantitative and qualitative, consistently illustrate that MAG-Edit outperforms existing methods in localized image editing within complex scenarios.
 We believe the proposed MAG-Edit scheme has pioneered a novel direction for applying localized editing in real-world scenarios.

 {
     \small   
     \bibliographystyle{ieeenat_fullname}
     \bibliography{main}
 }
\clearpage
\appendix
\appendixpage

\section{Summary}
In this appendix, we present more implementation details, additional experiments, and additional results as follows:
\begin{compactitem}
\item We present more implementation details of MAG-Edit in \secref{implementation}.
Furthermore, \secref{baselines} illustrates more implementation details on the benchmark dataset, baselines, quantitative metrics, and user study.
\item In \secref{compare}, we extend our comparisons to encompass training and fine-tuning methods, as well as recent developments in inversion techniques.
For a more comprehensive understanding of our proposed method, additional ablation studies are conducted and detailed in \secref{supp-ablations}.

\item We
demonstrate additional qualitative results to complement the paper in \secref{results}.

\item Finally, the limitations of our approach are thoroughly analyzed in \secref{limitations}.

\end{compactitem}

\section{Implementation Details}
\label{sec:implementation}
%
We utilize the official pre-trained Stable Diffusion v1.4 model\footnote{\url{https://github.com/CompVis/stable-diffusion}} as our foundation model.
The denoising sampling process employs the DDIM method~\cite{song2020denoising} over $T=50$ steps, maintaining a constant classifier-free guidance scale of $7.5$. 
 CA injection is performed during $[T, \tau_1]$.
For varying editing requirements, we set $\tau_1 = 10$ for color and texture edits, and $\tau_1 = 40$ for shape variation edits.
Our \textbf{MAG-Edit} optimization takes place during diffusion steps in the range $[T,\tau_2]$, with $\tau_2$ empirically set to $25$. 
For the gradient guidance process, we follow~\cite{chen2023training} by setting the gradient update scale $\delta$ using a linear scheduling rate as $\sqrt{(1-\alpha_t)/\alpha_t}$, particularly to optimize the token ratio constraint $\mathcal{L}_{TR}$.
This approach modulates the gradient's magnitude based on the denoising progress. 
On the contrary, for the constraint of the spatial ratio $\mathcal{L}_{SR}$, we keep $\delta=1$.
The optimization process is also influenced by the maximum number of iterations, empirically set $\rm{MAX\_IT}=15$. 
In cases involving negative prompt constraints, we empirically set $\lambda_p=2.5$ and $\lambda_{ng}=5.5$. 
%
To further preserve the structure of the original image, we also consider incorporating self-attention as P2P~\cite{hertz2022prompt} and replace them at diffusion steps $t\in [T,25]$.
Towards the end of the denoising process $t \in [15,0]$, we implement a latent blend operation from P2P~\cite{hertz2022prompt} to maintain information outside the edited region mask $\mathcal{M}$.
When evaluated on an Nvidia A100 (40GB) GPU, the runtime of MAG-Edit is around $1\thicksim 5$ minutes, varying with the selected values of $\rm{MAX\_IT}$ and $\tau_2$.

\section{Details of Comparisons with Baselines}
\label{sec:baselines}
\subsection{Benchmark Dataset}
\label{sec:benchmark}
\begin{figure*}[!t]
    \centering
    \includegraphics[width=0.99\linewidth]{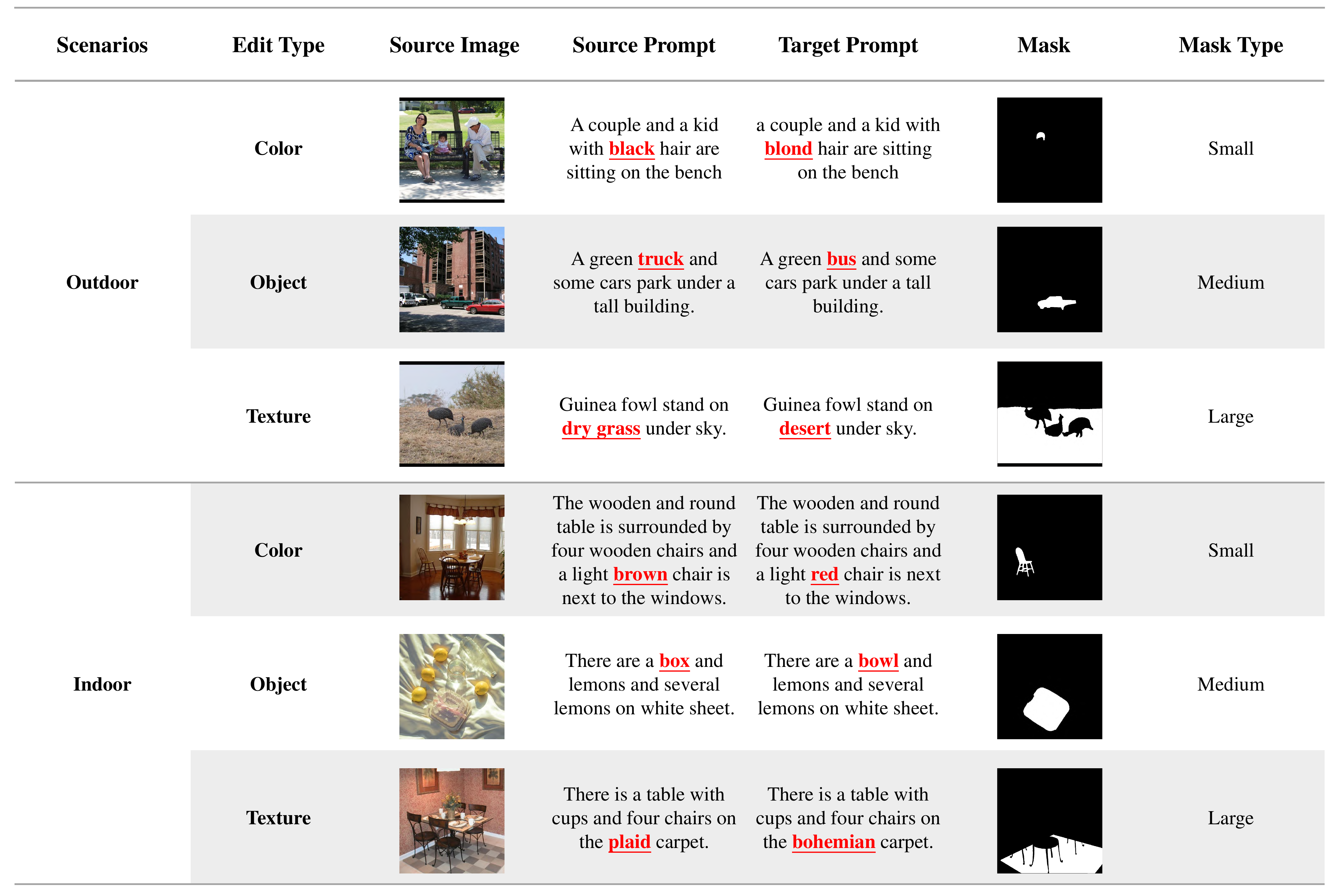}    
    \caption{\textbf{Examples images and annotations in the MAG-Bench dataset.
} 
}
\vspace{-3 mm}
    \label{fig:datasets}
\end{figure*}

Current datasets for text-based image editing methods are primarily limited to simple scenes with prominent objects. To enable a more thorough evaluation of our method, we have developed a benchmark dataset, named MAG-Bench, consisting of 200 images sourced from ADE20K~\cite{zhou2017scene}, MS-COCO~\cite{lin2014microsoft}, Cityscape~\cite{cordts2016cityscapes}, and the Internet. 
This dataset features complex scenes with multiple objects in various real-world indoor and outdoor settings, encompassing a wide range of object categories like humans, furniture, animals, vehicles, and food.
MAG-Bench is specifically designed to assess three types of local editing: (1) color editing, (2) texture editing which includes changes in material, background, and style, and (3) object replacement.
For the generation of source and target prompts, we initially utilized GPT-4~\cite{OpenAI2023GPT4TR}, followed by manual refinement to ensure the accuracy and relevance of these prompts.
The corresponding editing masks for each image are derived using the Segment Anything method\footnote{\url{https://github.com/facebookresearch/segment-anything}}. 
Acknowledging the critical role of the mask's size in localized editing, we initially classify each image into three categories based on mask size: relatively small, medium, and relatively large.
We then ensure a balanced distribution of varying sizes of editing regions across the datasets.
Thus, each image in MAG-Bench is accompanied by three annotations: a source prompt, a target edit prompt, and an edit region mask, as illustrated in \figref{datasets}.

\subsection{Implementation Details of Baselines}
\label{sec:baselineimplementation}
We use the official codes released by the authors for Blended LD\footnote{\url{https://github.com/omriav/blended-latent-diffusion}}, P2P\footnote{\url{https://github.com/google/prompt-to-prompt}}, and PnP\footnote{\url{https://github.com/MichalGeyer/plug-and-play}}. 
For DiffEdit~\cite{couairon2022diffedit}, we adopt the implementation from InstructEdit\footnote{\url{https://github.com/QianWangX/InstructEdit}}, which enhances automatic mask generation for scenarios involving multiple objects. This implementation, while improving upon mask generation, does not modify the core editing algorithm of DiffEdit~\cite{couairon2022diffedit}.
To facilitate fair comparisons, all methods use \emph{identical masks} provided in our benchmark dataset.
Notably, for DiffEdit~\cite{couairon2022diffedit} and P2P~\cite{hertz2022prompt}, we utilize ground-truth masks instead of those generated through unsupervised learning or derived from average CA maps.
In the case of P2P~\cite{hertz2022prompt}, we also integrate Null-text inversion~\cite{mokady2023null} as our approach for encoding real images.
With the exception of Blended LD~\cite{avrahami2023blended}, which solely focuses on the target edit description for the foreground region and omits tokens for other unedited areas, all other methods employ target prompts identical to those used in our method.

\subsection{Evaluation Details}
\label{sec:evaluationdetails}
We utilize the CLIP score with the CLIP ViT-L/14 model, as implemented in\footnote{\url{https://github.com/showlab/loveu-tgve-2023}}, and the DINO-ViT self-similarity distance, available at\footnote{\url{https://github.com/omerbt/Splice}}, as our evaluation metrics. 
To precisely evaluate localized editing, we crop the editing regions in both the source and edited images using bounding boxes as~\cite{huang2023pfb}.
This approach enables us to specifically assess text prompt alignment within these localized regions by calculating the CLIP score on the target edited tokens with the respective cropped edited image.
For instance, in a scenario where the editing objective is to alter a car's color to red, the CLIP score is computed using the phrase ``red car.''
This calculation excludes common tokens shared between the source and target prompts and focuses solely on the cropped image depicting the edited car and the target phrase.
To evaluate structure preservation within the localized editing regions, we utilize the DINO-ViT self-similarity by calculating the distance between the cropped source image and the corresponding cropped edited image.
\begin{figure*}[!t]
    \centering   \includegraphics[width=0.92\linewidth]{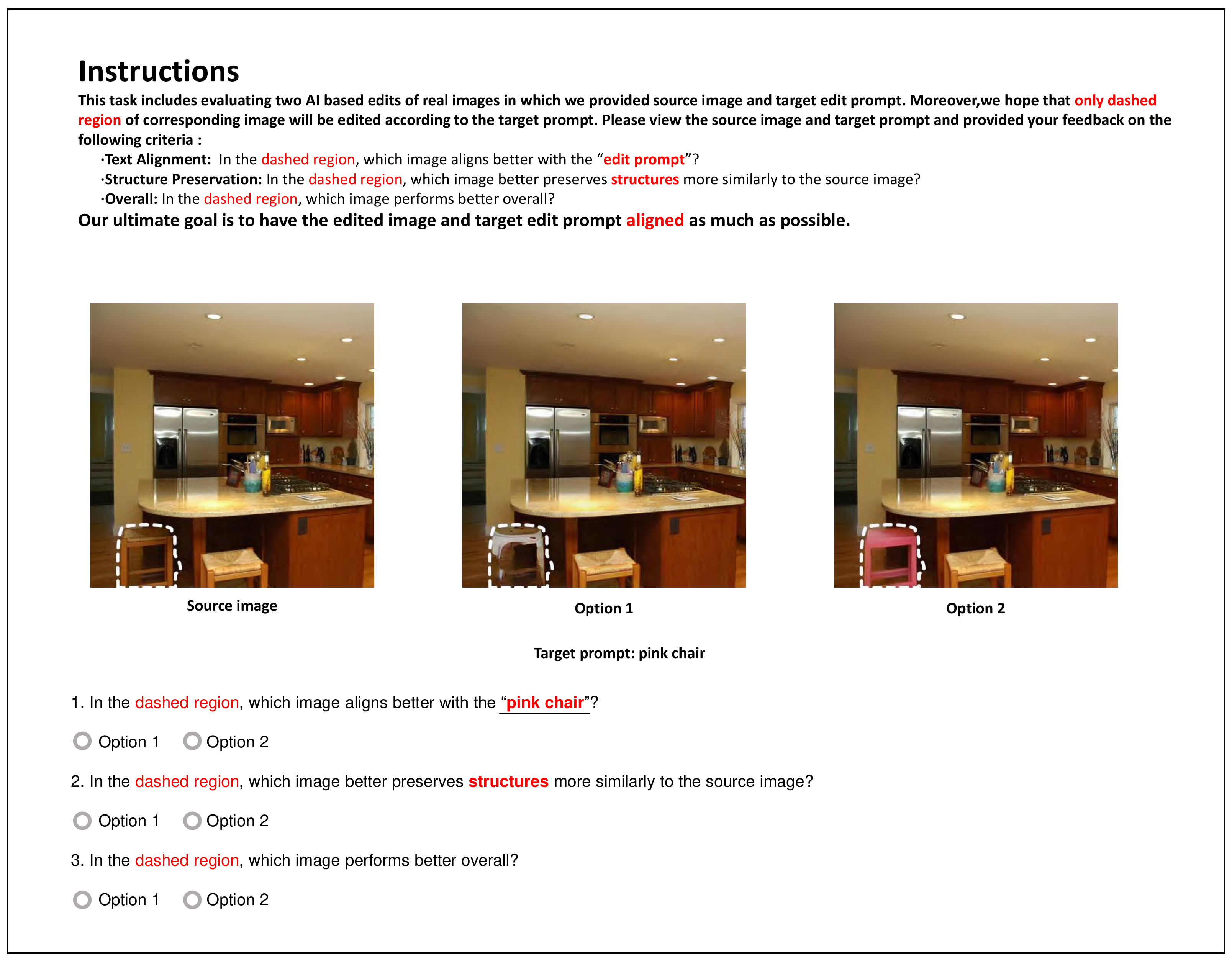}  
    \caption{\textbf{Example of one task for 5 human raters on Amazon MTurk to complete.
} }
\vspace{-3 mm}
    \label{fig:user-study}
\end{figure*}

%
\subsection{Details of User Study}
\label{sec:userdtudydetails}
We conduct a user study on the Amazon MTurk platform\footnote{\url{https://requester.mturk.com}}.
The user study comprises over 120 tasks, each evaluated by five human evaluators, as depicted in \figref{user-study}.
In each task, participants are presented with a source image alongside two edited images: one generated by our proposed method and the other by a randomly selected baseline method, with their presentation order shuffled.
To enhance the visibility of localized editing regions, we outline the prospective edit regions with white dashed lines in each pair of comparison images and their corresponding source images, as illustrated in \figref{user-study}.
Additionally, a simplified version of the target edit prompt was displayed beneath the comparison images.
We then pose three questions for the raters to answer: 
\begin{compactitem}
\item Text Alignment: In the dashed region, which image aligns better with the ``edit prompt''? 
\item Structure Preservation: In the dashed region, which image preserves structures more similarly to the source image? 
\item Overall: In the dashed region, which image performs better overall?
\end{compactitem}
To ensure the credibility and reliability of our user study, we only involve Amazon MTurk workers with `Master' status and a Human Intelligence Task (HIT) Approval Rate exceeding $90\%$ across all Requesters' HITs.
In total, the 120 tasks garnered responses from 600 distinct human evaluators.
%
%
\begin{figure}[!t]
    \centering
    \includegraphics[width=1\linewidth]{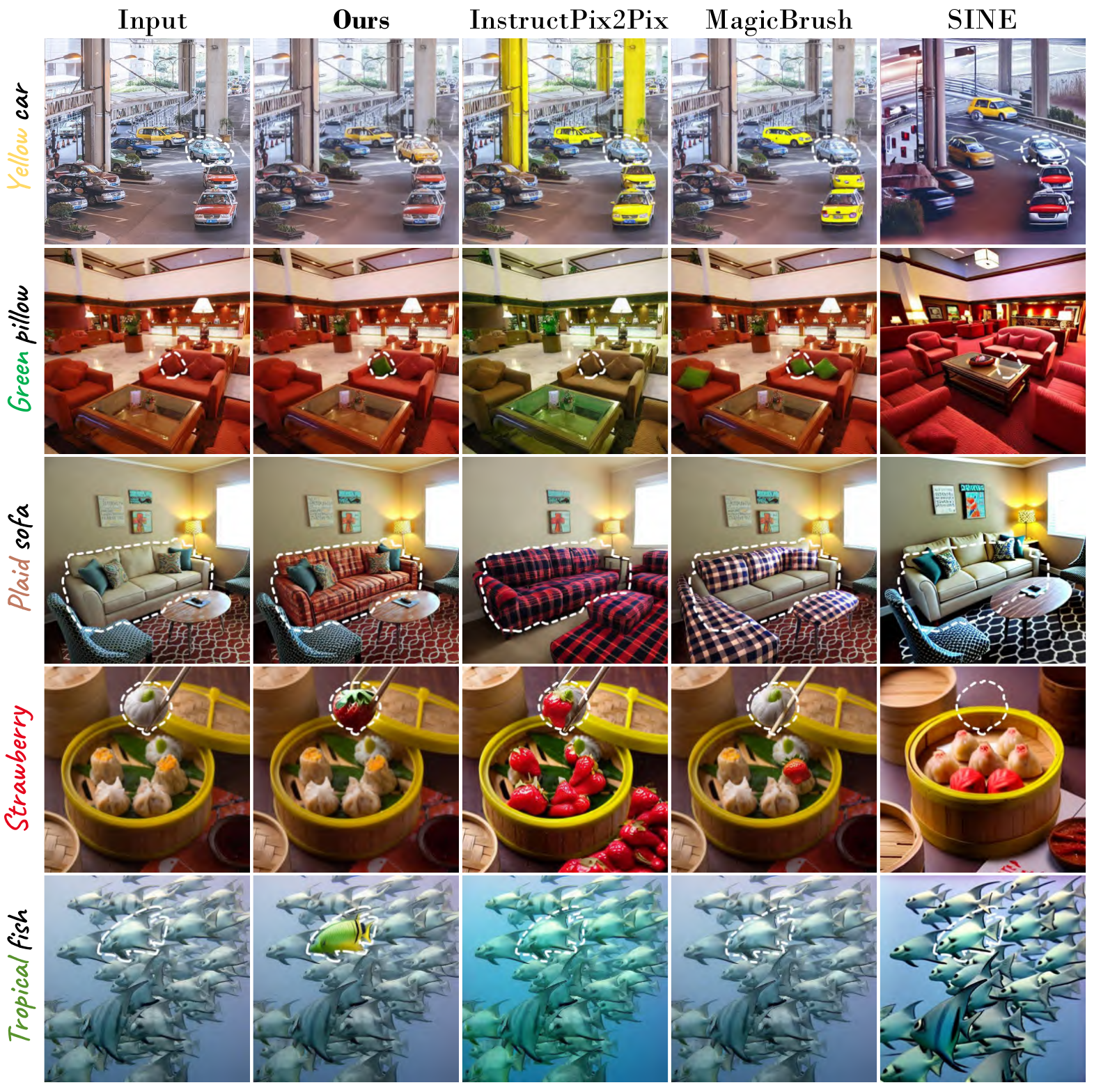}    \caption{\textbf{Qualitative comparisons with training and fine-tuning methods for localized editing in complex scenarios.} 
    Training approaches such as InstructPix2Pix~\cite{brooks2023instructpix2pix} and MagicBrush~\cite{zhang2023magicbrush} demonstrate issues like leakage or unintended modifications in structure. The fine-tuning method SINE~\cite{zhang2023sine} is ineffective in both reconstructing and generating desired editing effects.
    }
    %
    \label{fig:compare}
\end{figure}

\section{Comparisons with Other Baselines}
\label{sec:compare}
In this section, we begin by comparing our approach with current training and fine-tuning methods, aiming to further validate the efficacy of our proposed method in facilitating localized editing in complex scenarios.
Subsequently, we extend our comparison to include recent advancements in training-free inversion methods. This comparison is intended to illustrate that despite improvements in inversion methods, they still face challenges in addressing localized editing issues.
%

\begin{figure}[!t]
    \centering
    \includegraphics[width=1\linewidth]{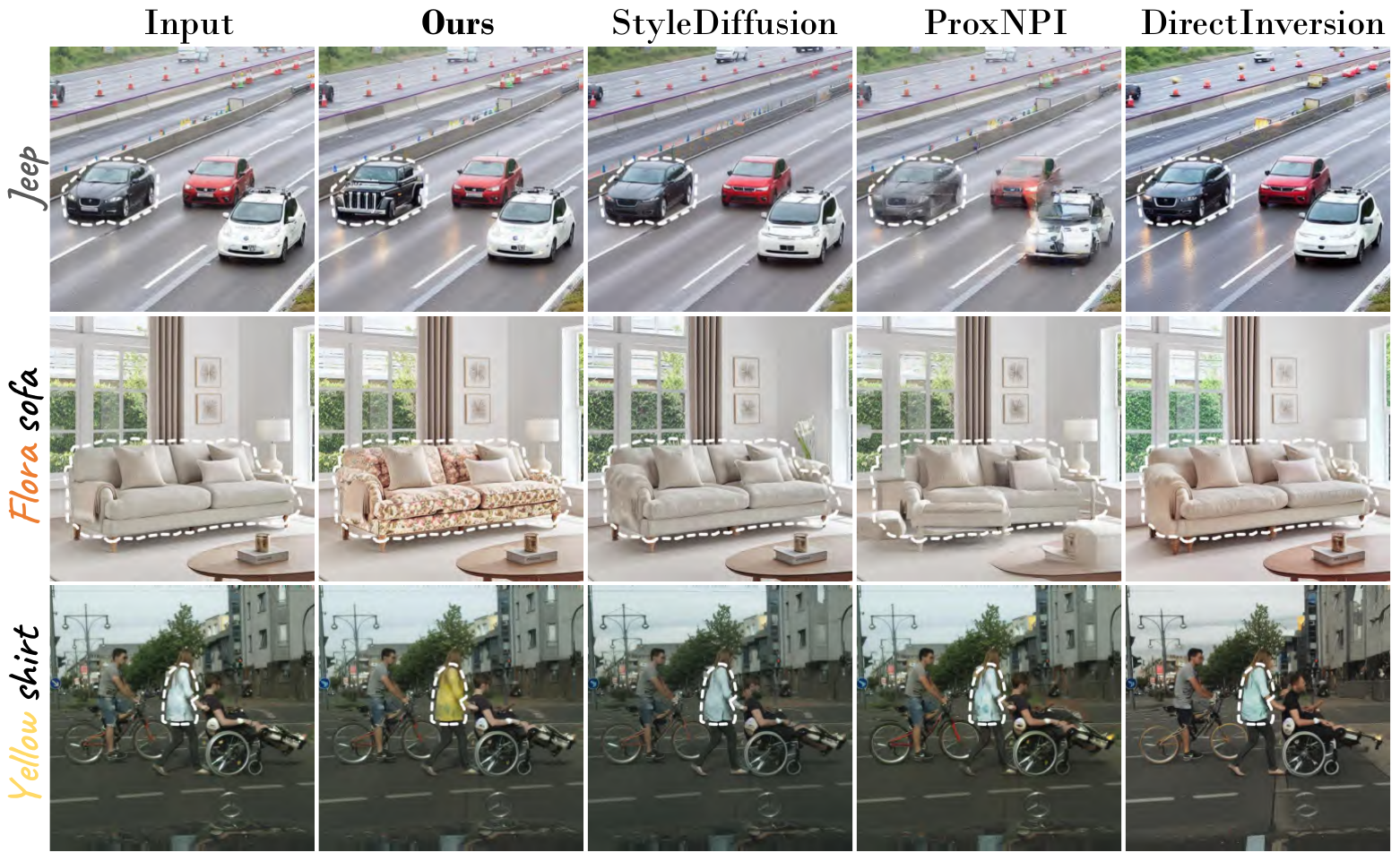}    \caption{\textbf{Qualitative comparisons with recent inversion methods for localized editing in complex scenarios.} 
Despite recent advancements, solely enhancing inversion methods continues to be inadequate for effective editing of localized regions in complex scenarios.
    }
    \vspace{-3 mm}
    \label{fig:compare-inv}
\end{figure}
\begin{figure}[!t]
    \centering \includegraphics[width=1\linewidth]{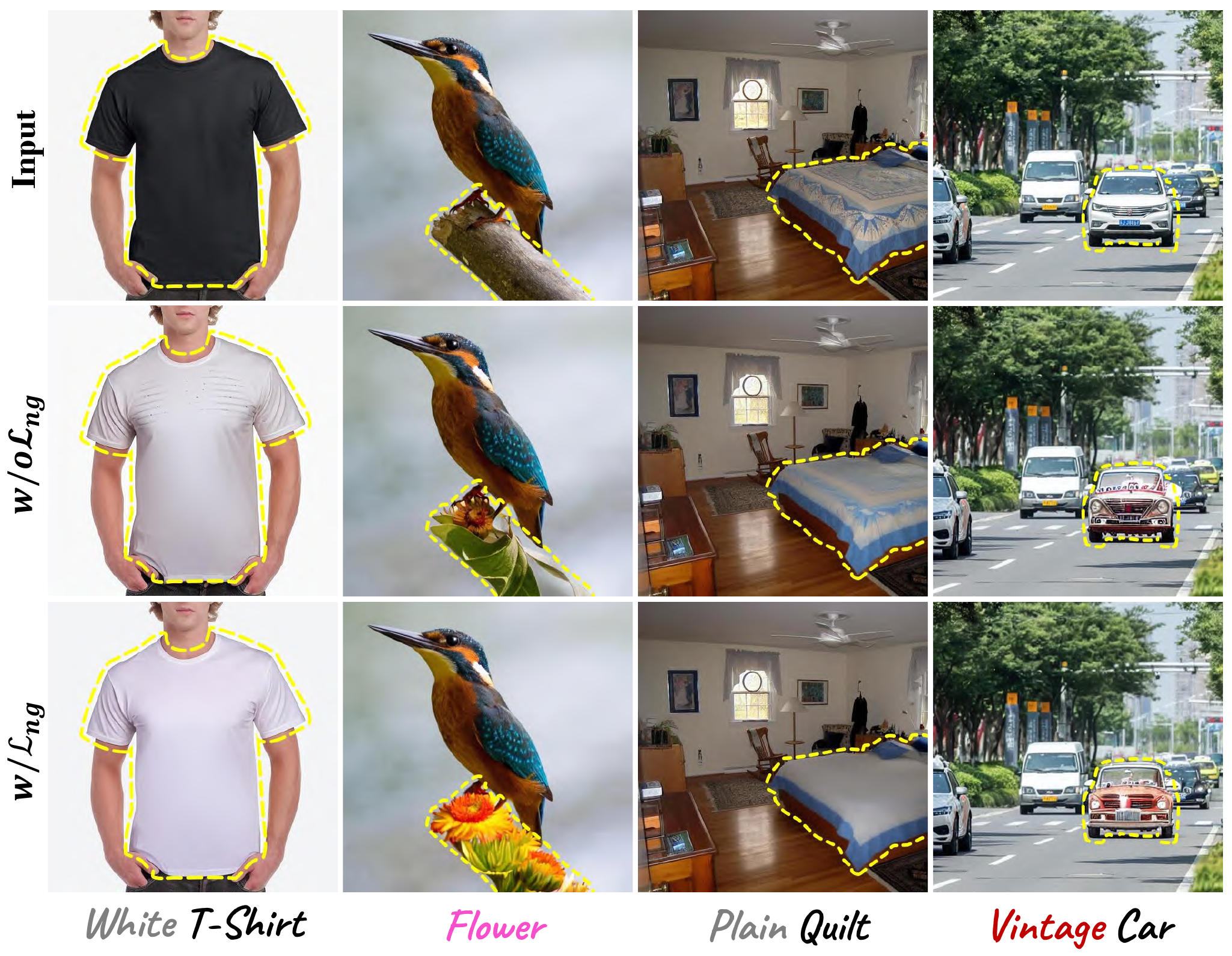}    \caption{\textbf{Ablation study on the negative prompt constraint.} 
Negative prompt
constraints can amplify the effectiveness of editing by diminishing
the influence of information from the original image.
    }
    \label{fig:ablation-ng}
    \vspace{-1 mm}
\end{figure}

\begin{figure}[!t]
    \centering \includegraphics[width=1\linewidth]{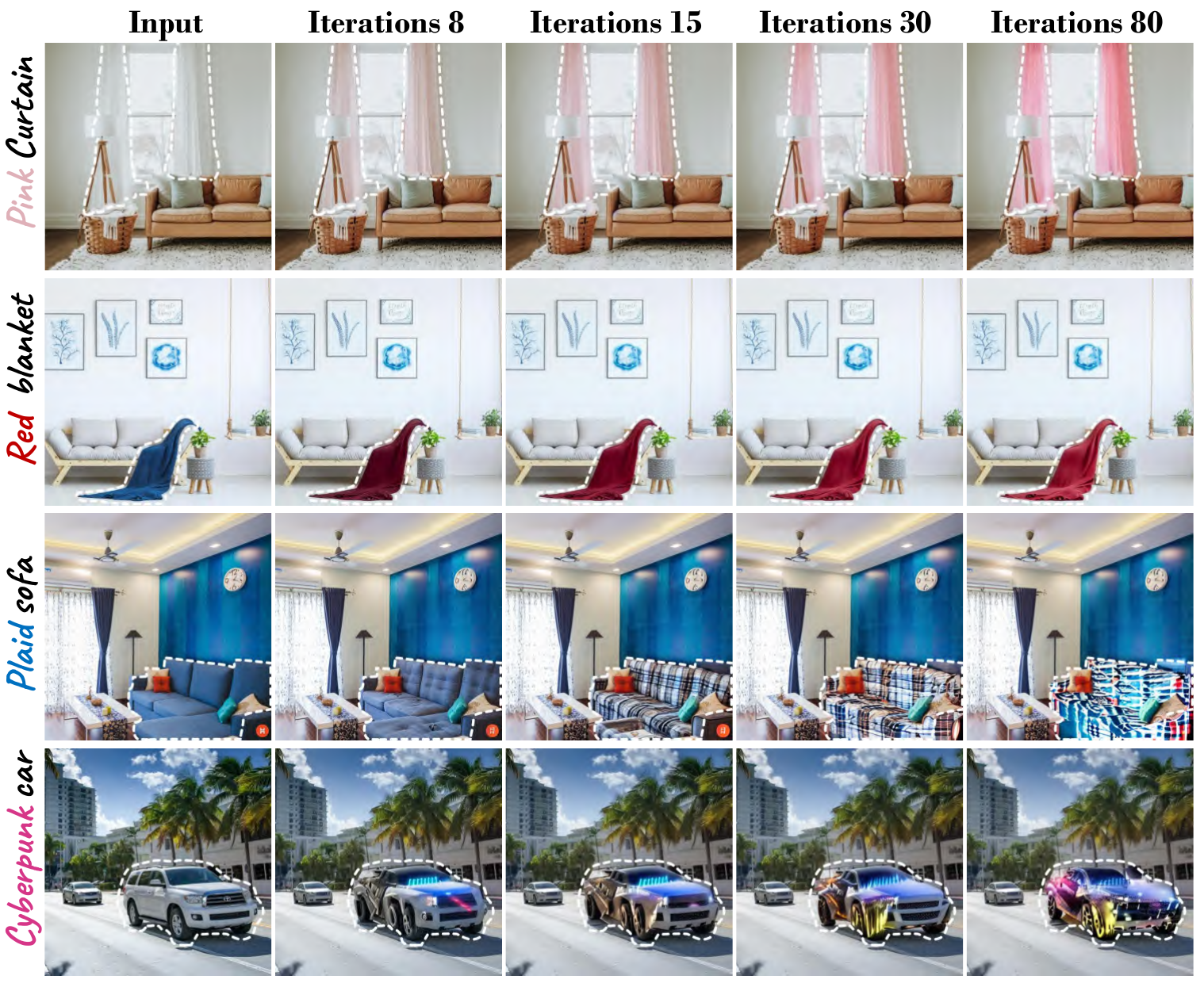}    \caption{\textbf{Impact of optimization iterations.} 
   Increasing the number of iterations enhances the granularity of editing. However, overly extensive iterations can lead to notable artifacts arising from structural modifications.
    }
    \vspace{-3 mm}
    \label{fig:ablation-iters}
\end{figure}

\begin{figure*}[!t]
    \centering \includegraphics[width=0.99\linewidth]{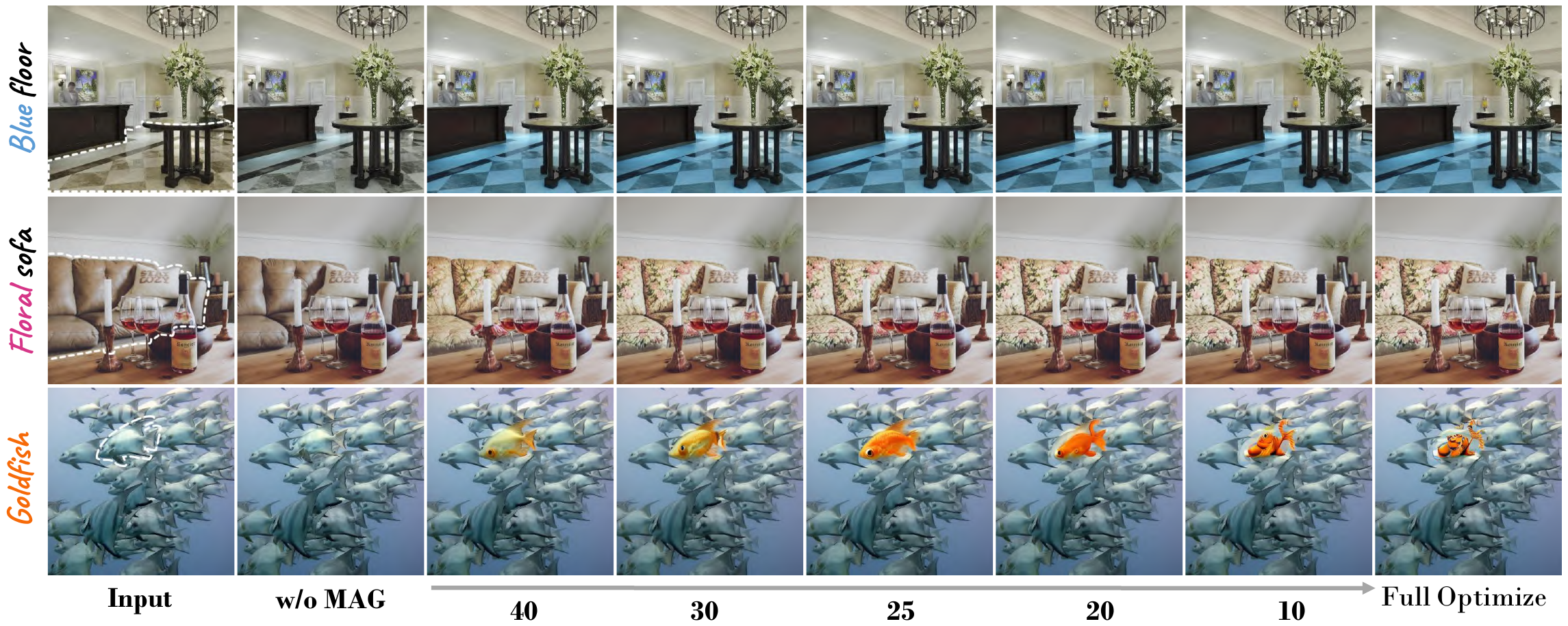}    
\caption{\textbf{Applying MAG-Edit through a varied number of diffusion steps.} 
    We use white dashed lines to demarcate the editing regions in the source images. 
    Each row demonstrates the optimization of the noise latent feature ranging from $0\%$ (left) to $100\%$ (right) of the steps.
    In particular, we assign values to $\tau_2=\{50,40,30,25,20,10,0\}$, indicating the end of the diffusion step range, from $50$ to $\tau_2$, as noted at the bottom of each image.
    Without MAG, there is a negligible localized editing effect in the intended regions.
    On the other hand,  employing MAG across all steps does not markedly enhance the granularity of color and texture editing. 
    Moreover, it results in noticeable structural artifacts in the shape editing.
    }
    \label{fig:ablation-steps}
\end{figure*}

\begin{figure*}[!t]
    \centering
    \includegraphics[width=0.95\linewidth]{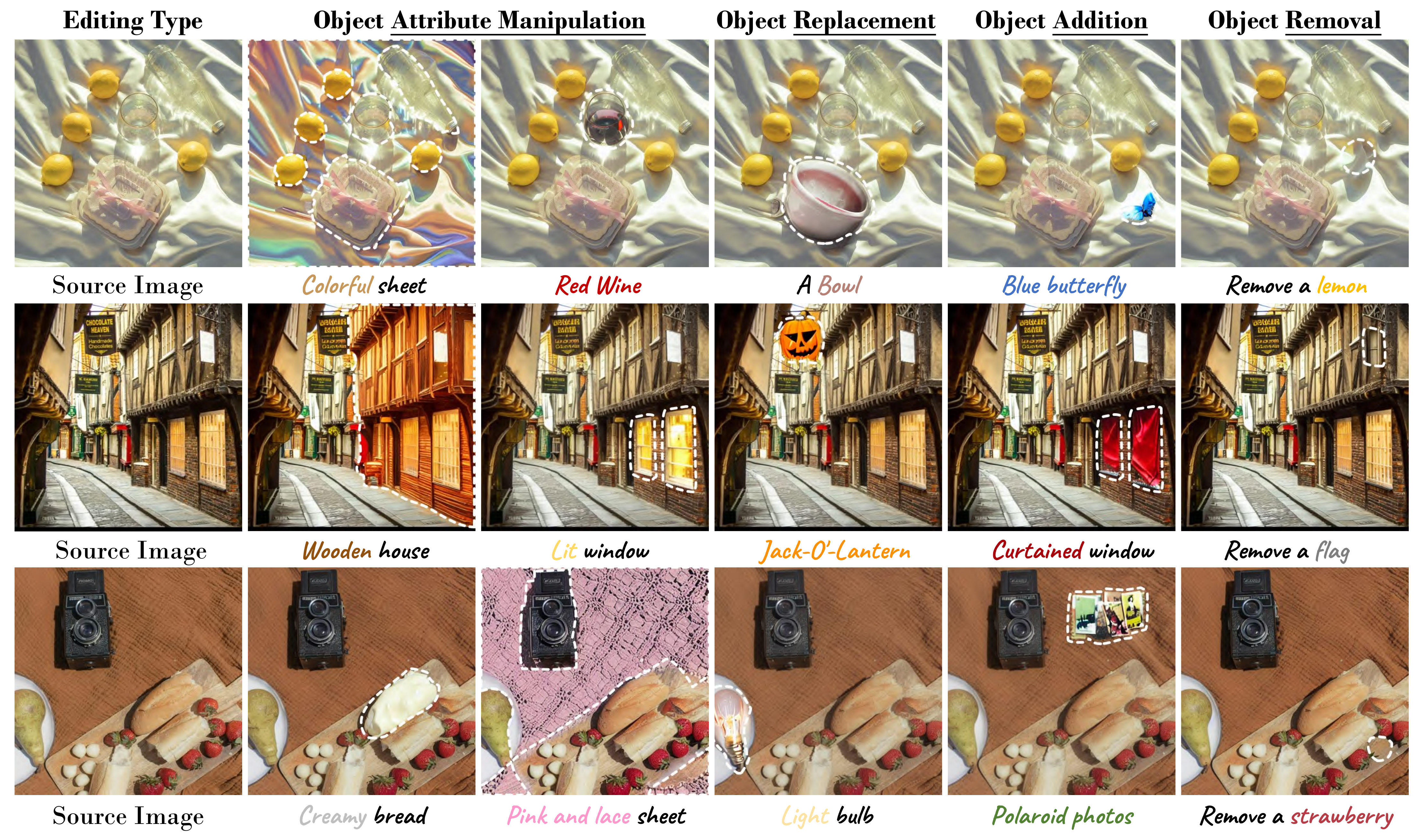}   \caption{\textbf{Various localized editing types.}
    We provide a simplified version of the corresponding target prompt under each edited image.
    }
    \label{fig:editing1}
     \vspace{-5 mm}
\end{figure*}

\begin{figure*}[!t]
    \centering
    \includegraphics[width=0.95\linewidth]{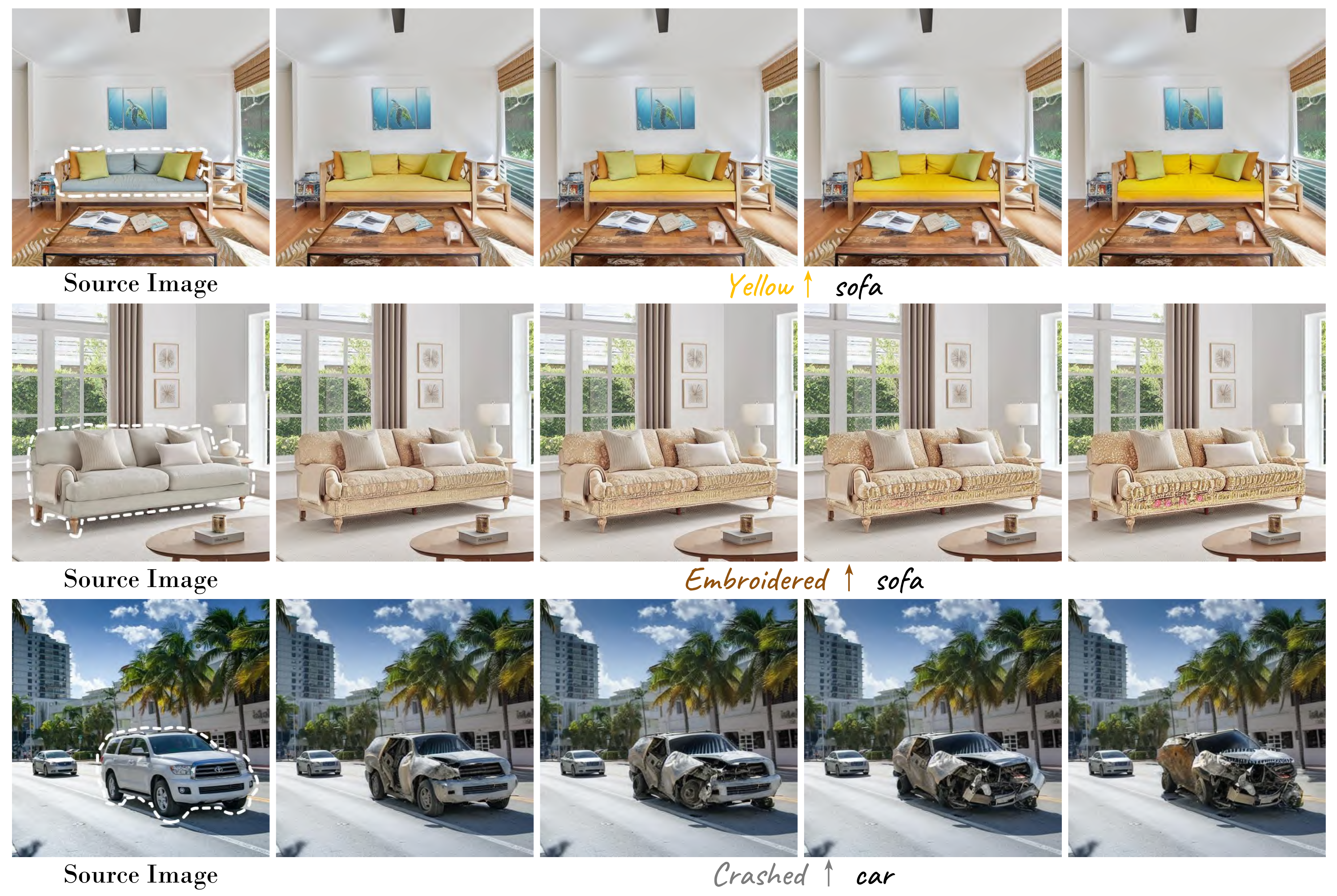}    \caption{\textbf{Granularity controllable localized editing.}
    We present a simplified version of the corresponding target prompt under the edited images.
$\uparrow$ denotes increasing the editing magnitude. }
    \label{fig:granularity}
       
\end{figure*}

\begin{figure*}[!t]
    \centering
    \includegraphics[width=0.995\linewidth]{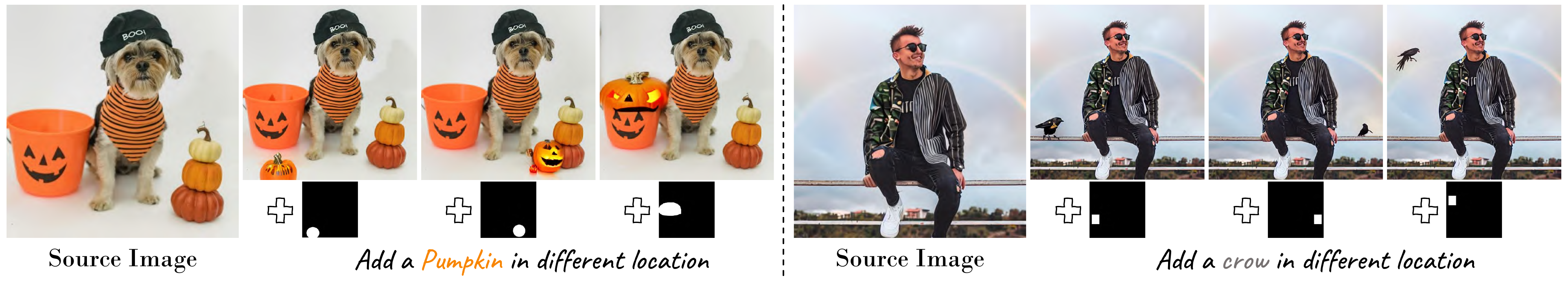}    \caption{\textbf{ Spatial controllable localized editing.} 
    }
    \vspace{-3 mm}
    \label{fig:spatial-control}
\end{figure*}

\subsection{Comparisons with Training and Fine-tuning Methods}
\label{sec:Fine-tuning}
We initiate our qualitative comparison with existing training methods by evaluating InstructPix2Pix~\cite{brooks2023instructpix2pix} and MagicBrush~\cite{zhang2023magicbrush}, utilizing their officially released codes and models.
InstructPix2Pix~\cite{brooks2023instructpix2pix} is trained on an extensive data set, which includes instructions generated by GPT-3 and image examples modified by P2P~\cite{hertz2022prompt}. This training facilitates instruction-based image editing during the inference phase.
MagicBrush~\cite{zhang2023magicbrush} harnesses a large-scale dataset of manually annotated real image editing triplets and optimizes the InstructPix2Pix model to improve editing capabilities.
For our comparisons, we utilize editing instructions such as ``make'' and ``change'' to manipulate images.
\figref{compare} illustrates that InstructPix2Pix, due to its lack of mask integration, frequently leads to substantial leakage into incorrect regions during localized editing in complex scenes.
In contrast, MagicBrush demonstrates better localized editing in some cases, thanks to mask-integrated examples in its dataset.
However, MagicBrush encounters difficulties in precisely localizing individual objects within scenes containing multiple similar objects. This challenge is evident in the first and second rows of \figref{compare}, where it struggles with tasks like coloring one car yellow and one pillow green.
Moreover, as shown in the third row of \figref{compare}, MagicBrush~\cite{zhang2023magicbrush} tends to modify the underlying structure in areas undergoing texture changes.
In contrast, our training-free method efficiently attains desired editing effects in the target local regions while preserving the original structure. 
A significant advantage of our approach is the elimination of the need for extensive training on large datasets, saving significant time and resources.

Subsequently, we compare our method with the existing fine-tuning method, SINE~\cite{zhang2023sine}, using the code provided by its authors.
SINE~\cite{zhang2023sine} proposes fine-tuning a pre-trained text-to-image (T2I) model with a single real image, incorporating model-based classifier guidance and patch-based guidance to prevent overfitting. 
However, as illustrated in \figref{compare}, SINE fails to generate any noticeable editing effects in the intended regions. Furthermore, it faces difficulties in accurately reconstructing the original image in complex scenarios.


\subsection{Comparisons with Recent Inversion Methods}
\label{secInversion}
To demonstrate that localized editing challenges are not sufficiently addressed by mere advancements in inversion techniques, we compare our method with recent inversion methods. 
This includes Style Diffusion~\cite{li2023stylediffusion}, ProxNPI~\cite{han2023improving}, and DirectInversion~\cite{ju2023direct}, utilizing their official codes. 
Each of these approaches incorporates P2P~\cite{hertz2022prompt} to facilitate editing capabilities.
As depicted in \figref{compare-inv}, it is evident that these recent inversion methods are unable to produce effective editing results in localized regions within complex scenarios. 
 This underscores the heightened challenges faced in localized editing within intricate compositions compared to simpler settings.
In contrast, our method significantly improves localized editing by optimizing the noise latent feature through our specially designed MAG mechanism.

\begin{figure}[!t]
    \centering
    \includegraphics[width=1.0\linewidth]{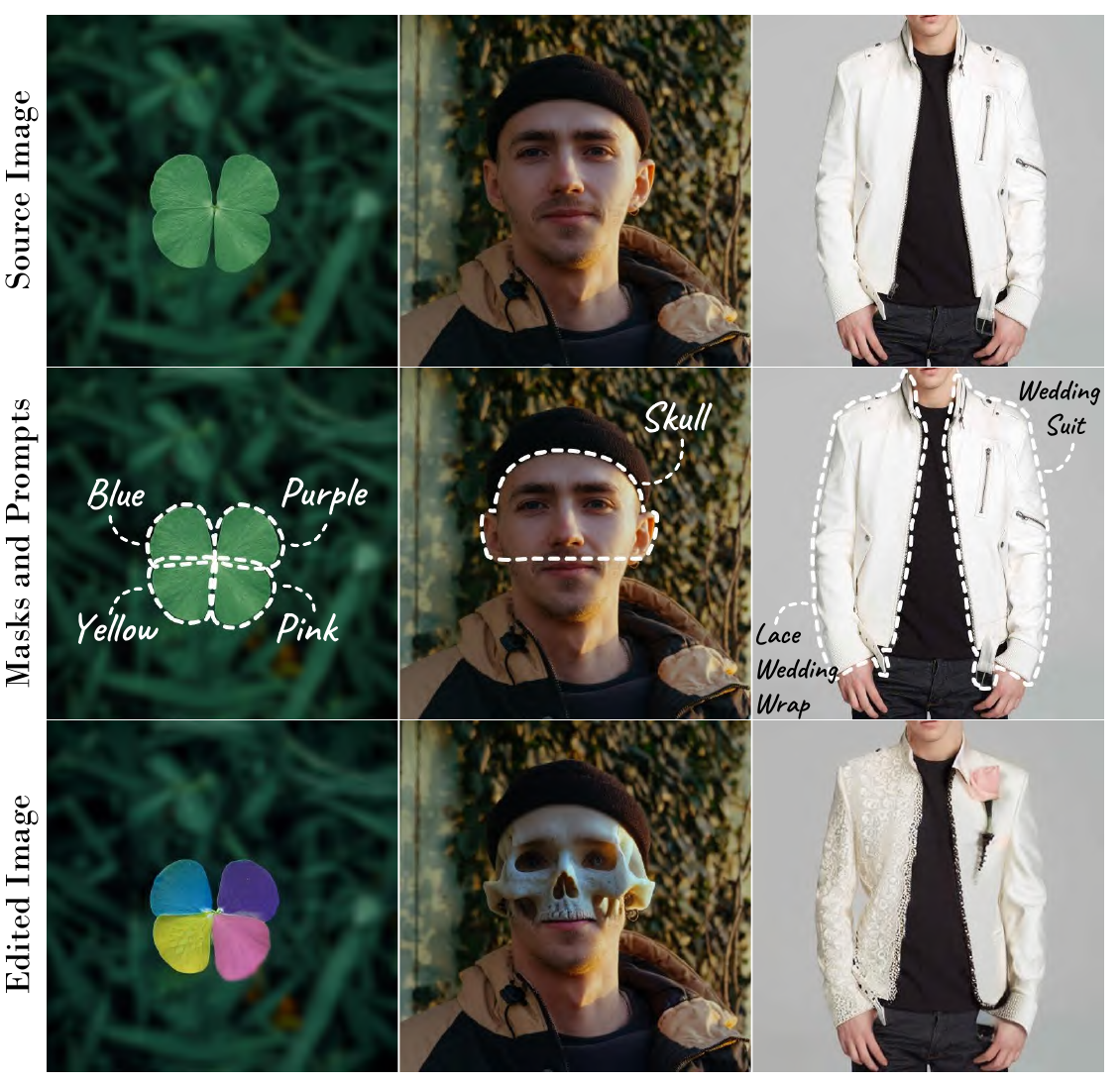}    \caption{\textbf{Part-level localized editing.}
In the second row of images, the editing regions are indicated with white dashed lines, and the target prompts are also annotated for clarity.
    }
    \vspace{-3 mm}
    \label{fig:part}
\end{figure}

\section{Additional Ablation Studies}
\label{sec:supp-ablations}
\Paragraph{Impact of Negative Prompt Guidance.} 
\figref{ablation-ng} demonstrates that negative prompt guidance is effective in diminishing the original image's information, which is beneficial when dealing with original images that have information significantly contrast with the target prompt.
For instance, as shown in the first column of \figref{ablation-ng}, when altering the color of a T-shirt from black to white, not applying negative constraints could lead to the edited image preserving some black elements. 
The negative prompt constraint, in such scenarios, efficiently reduces this residual black information.
Moreover, as observed in the third column of \figref{ablation-ng}, when transforming a patterned quilt into a plain one, the negative prompt constraint plays a crucial role in diminishing the original textures of the quilt.

\Paragraph{Impact of Optimization Iterations.}
The number of maximum iterations for optimizing the noise latent feature is crucial in modulating the magnitude of editing.
As shown in \figref{ablation-iters}, increasing the number of iterations can improve the granularity of the editing. However, in texture and shape editing, excessive iterations may result in significant artifacts as a result of alterations in the structure.

\Paragraph{Impact of Optimization Diffusion Steps.}
Applying MAG-Edit across various diffusion steps significantly impacts the final editing results. 
\figref{ablation-steps} demonstrates that optimization in the initial diffusion steps can quickly alter the color, indicating that optimization within the $t\in[T,40]$ steps is generally sufficient for color editing.
On the contrary, texture and shape edits necessitate a greater number of diffusion steps. 
Updating the latent noise feature after $25$ steps does not significantly improve texture editing granularity but requires extended optimization time.
In shape editing, over-optimization after $25$ steps can lead to pronounced artifacts, due to structural changes.

\begin{figure}[!t]
    \centering
    \includegraphics[width=1\linewidth]{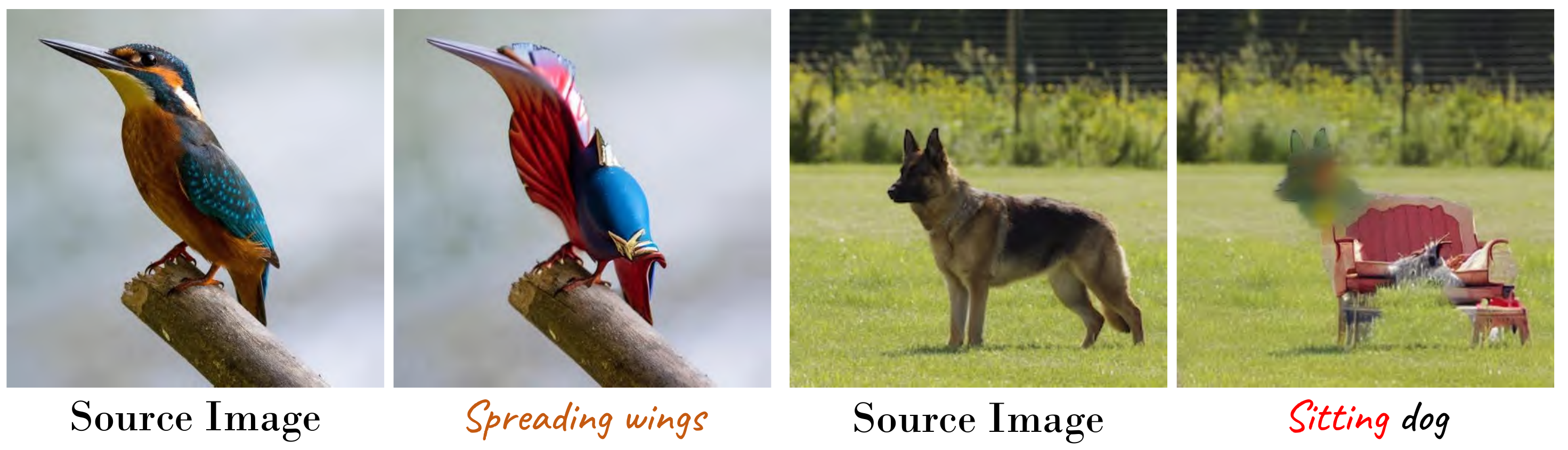}    \caption{\textbf{Editing failure cases.}
    Due to its reliance on maintaining the structure using the CA maps of the reconstruction branch, the proposed method encounters limitations in editing images that necessitate significant pose alterations. 
     For example, changing the dog from ``standing'' to ``sitting''.
    }
        \vspace{-3 mm}
    \label{fig:limitation}
\end{figure}
\section{Additional Results}
\label{sec:results}

\begin{figure*}[!t]
    \centering
    \includegraphics[width=0.98\linewidth]{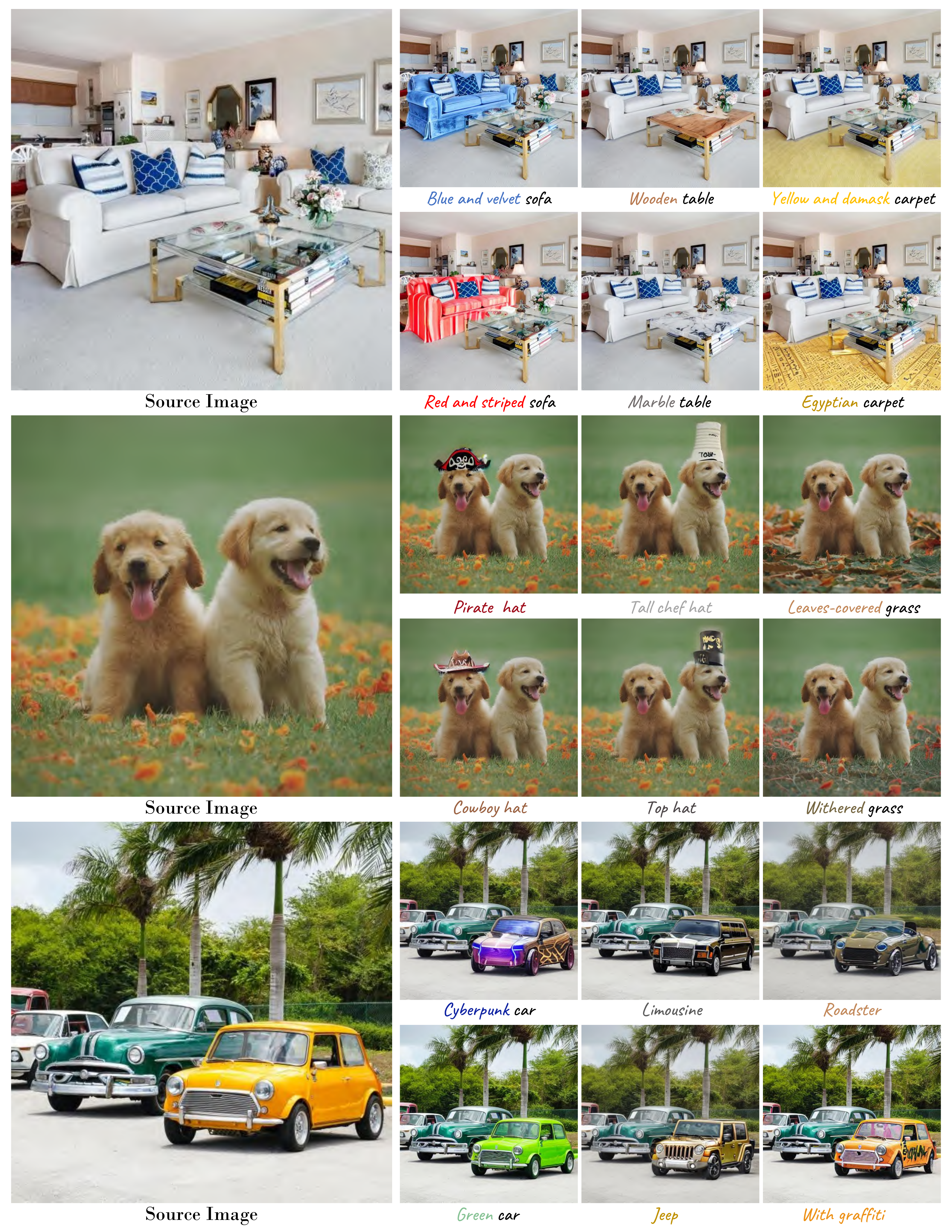}    \caption{\textbf{Various localized editing types.}
In each edited image, we present a simplified version of the corresponding target prompt.
    }
    \label{fig:editing2}
\end{figure*}

\begin{figure*}[!t]
    \centering
    \includegraphics[width=0.995\linewidth]{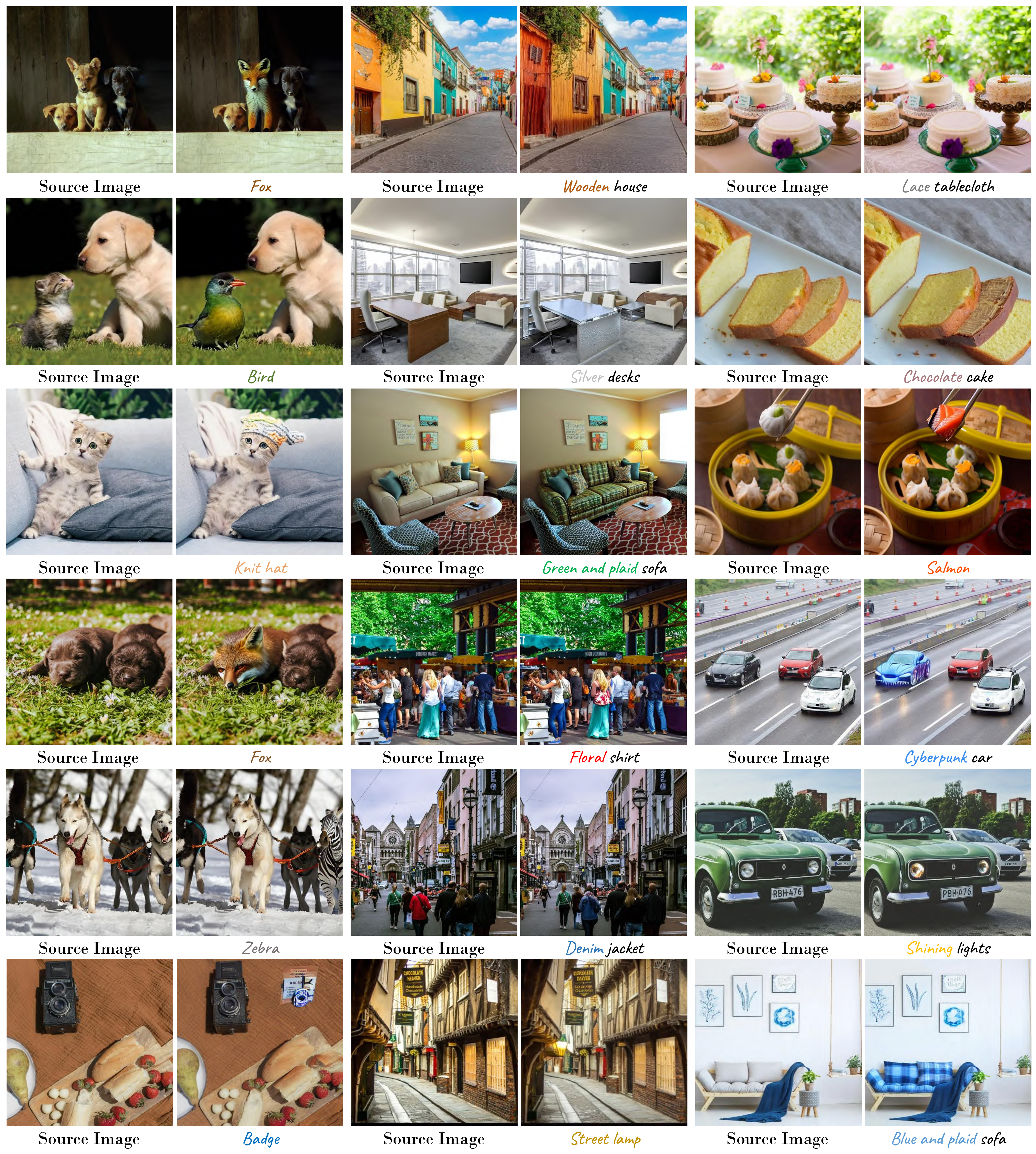}    \caption{\textbf{Additional results on localized editing in complex scenarios.}
   We provide a simplified version of the target prompt beneath each edited image.
    }
    \vspace{-5 mm}
    \label{fig:editing3}
\end{figure*}

Our method offers a broad spectrum of localized editing capabilities, encompassing object attribute manipulation (\eg color and texture), object replacement, insertion, and removal, as exemplified in \figref{editing1}. 
Additional examples of localized editing in complex scenarios are illustrated in \figref{editing2} and \figref{editing3}.
Furthermore, we demonstrate the controllability of our localized editing approach in terms of both the magnitude of edits in \figref{granularity} and their spatial precision in \figref{spatial-control}. 
This allows for precise adjustment of editing granularity and the application of editing effects in various locations, catering to a variety of user requirements.

A key advancement of our method is \emph{its extension from object-level to more intricate part-level localized editing}, thereby enabling the integration of various editing effects within distinct parts of a single object. As demonstrated in \figref{part}, our method is capable of sophisticated editing, such as altering a four-leaf clover into a four-colored flower, or the creation of garments with mixed textures, showcasing its versatility and precision in fine-grained localized editing tasks.

\section{Limitations and Future Work}
\label{sec:limitations}
The MAG-Edit method has shown effective capabilities in localizing edits within complex scenarios, but it also has its limitations, which are a key focus of our future research efforts.
A primary limitation is the method's inference time attributed to the optimization process, which takes around $1\thicksim 5$ minutes on an A100 GPU to edit a single image. Future work will focus on developing strategies to accelerate this optimization process.
Furthermore, our method relies heavily on maintaining structure through CA maps in the reconstruction branch. However, it falls short in editing tasks that demand substantial pose changes. 
As illustrated in \figref{limitation}, an example of this limitation is observed in the task of transitioning a standing dog to a seated position, as discussed in~\cite{cao_2023_masactrl}. We acknowledge this challenge and plan to explore solutions in future work, potentially involving adjustments in how the SA is injected from the reconstruction branch to the editing branch.

\end{document}